\documentclass{article}
\usepackage{iclr2027_conference,times}
\usepackage[T1]{fontenc}
\usepackage[utf8]{inputenc}
\usepackage{amsmath,amssymb,booktabs,multirow,graphicx,microtype,xcolor,fvextra,textcomp}
\usepackage{wrapfig}
\usepackage{caption}
\usepackage{algorithm,algpseudocode,float}
\usepackage{url}

\iclrfinalcopy
\usepackage[
    pagebackref,
    breaklinks,
    citecolor=uclablue,
    linkcolor=uclablue,
    colorlinks=true,
]{hyperref}
\definecolor{uclablue}{rgb}{0.15, 0.45, 0.68}

\newcommand{\method}{UniEvo-VL}
\newcommand{\sg}{\operatorname{sg}}
\newcommand{\E}{\mathbb{E}}
\newcommand{\ours}{\textbf{UniEvo-VL}}
\title{UniEvo-VL: An On-policy Self-Distillation Training Recipe for Multimodal Model Self-improvement}

\author{
\textbf{Fang Wu}$^{\heartsuit}$\thanks{Equal contribution.},\,
\textbf{Da Xing}$^{\dagger}$\footnotemark[1],\,
\textbf{Ji Wang}$^{\spadesuit}$\footnotemark[1], 
\textbf{Yanjie Huang}$^{\circ}$\footnotemark[1],\,
\textbf{Junxi Wang}$^{\circ}$\, \\
\textbf{Hejia Geng}$^{\clubsuit}$,\,
\textbf{Guancheng Wan}$^{\circ}$,\,
\textbf{Bowen Zuo}$^{\diamondsuit}$,\,
\textbf{Xiaomin Li}$^{\square}$,\,
\textbf{Shixiang Tang}$^{\circ}$ \\
\textbf{Xinyu Xiang}$^{\heartsuit}$,\,
\textbf{Zehong Wang}$^{\triangle}$,\,
\textbf{Shiyi Du}$^{\nabla}$,\,
\textbf{Peng Xia}$^{\star}$,\,
\textbf{Shuangjia Zheng}$^{\circ}$ \\
\textbf{Yining Hong}$^{\heartsuit}$,\,
\textbf{Li Erran Li}$^{\circ}$,\,
\textbf{Jure Leskovec}$^{\heartsuit}$,\,
\textbf{Yejin Choi}$^{\heartsuit}$ \\
$^{\heartsuit}$Stanford University,\,
$^{\dagger}$Johns Hopkins University,\,
$^{\circ}$Independent Researcher \\
$^{\spadesuit}$University of Toronto,\,
$^{\clubsuit}$University of Oxford,\,
$^{\diamondsuit}$UC, Riverside 
$^{\square}$MatrAIx,\, \\ 
$^{\triangle}$University of Notre Dame,\,
$^{\nabla}$Carnegie Mellon University 
$^{\star}$UNC--Chapel Hill \\
}

\begin{document}
\maketitle
\begin{abstract}
Modern multimodal models bring generation and understanding into a single unified system, which enables them to provide and learn from their own feedback. Motivated by this unified capacity, we introduce UniEvo-VL, a self-evolving framework for multimodal models to learn from this constructive self-correction feedback during test-time compute. 
Instead of relying on a separately trained, stronger generator as teacher, we use critiques of the generator's own outputs, produced by a fixed vision-language critic, as privileged information. The generator and its EMA copy act as student and teacher under different contexts.
The student only sees the vanilla question, while the teacher conditions on the privileged critique. Then training minimizes the per-state divergence between their denoising diffusion distributions over the student's own sampling trajectories. 
Experiments demonstrate that UniEvo-VL improves the image generation capabilities of multimodal models, while maintaining their sensitivity to additional reflection information.
With Qwen-Image-2512, direct-generation performance improves from 0.747 to 0.808 on GenEval using Qwen feedback and from 32.97 to 35.53 on GenEval2 Soft-TIFA using the external Luna critic. 
Moreover, attempts with more powerful external critics (e.g., GPT5.6-Luna) show that multimodal models with strong judge capabilities can anticipate a higher self-evolving ceiling. Last but not least, mixed text-rendering outcomes show that our self-improvements may not be uniform across different tasks. Our study aims to shed light on the current hot recursive self-improvement research line to enhance the user experience when using multimodal models without paired ground-truth images, human annotation, or task-specific reward functions. 
\end{abstract}

\begin{figure}[H]
\centering
\includegraphics[width=\linewidth]{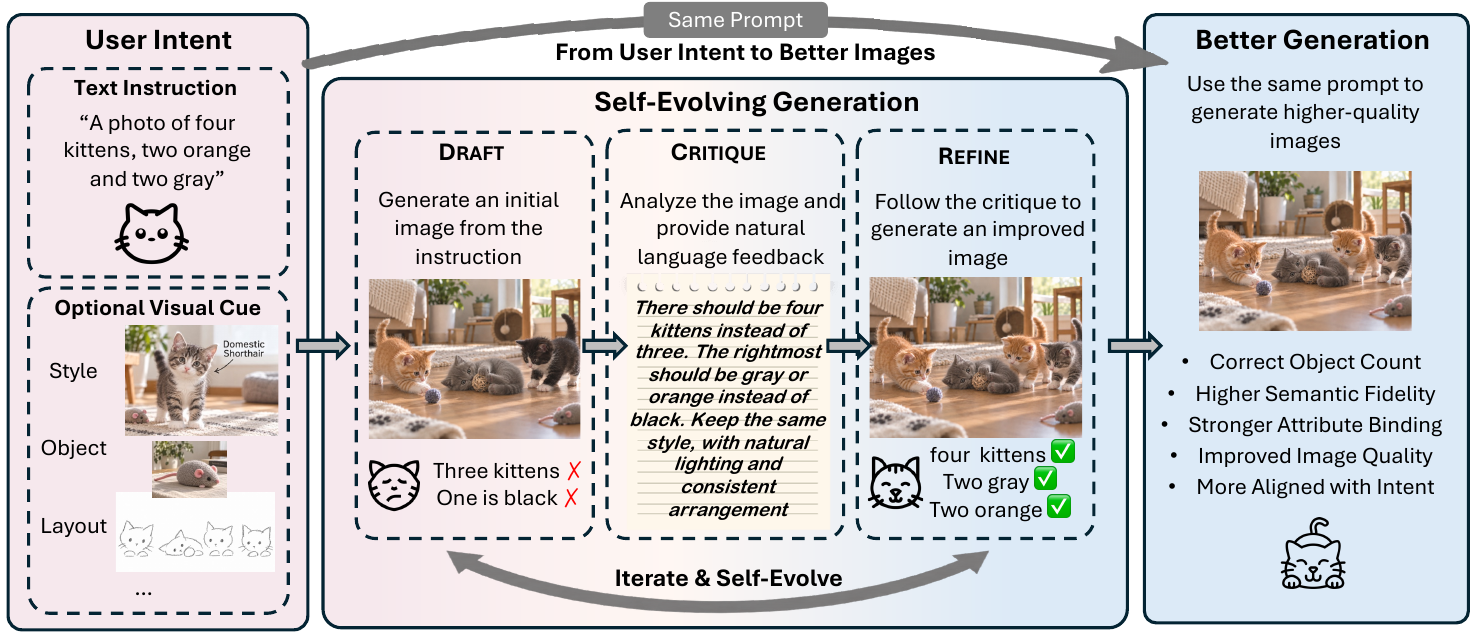}
\caption{\textbf{From understanding to self-evolution.} Generated drafts become training experience through visual critique and generator updates. The conceptual loop motivates both self-critique and an optional extension with a stronger external critic.}
\label{fig:concept}
\end{figure}

\section{Introduction}

Unified multimodal models (MMM)~\citep{achiam2023gpt,team2023gemini}  exhibit great capabilities in both visual generation and understanding. This creates an unprecedented opportunity for them to self-evolve without external supervision. In particular, MMMs can identify discrepancies between their generated images and the given instructions, and use this self-feedback for future improvements. Inspired by this relationship, prior work proposes to self-enhance MMMs by selecting self-generated outputs for supervised fine-tuning and preference optimization~\citep{internalgap}, or reconstructing self-generated interactions into captioning, judgment, and reflection training tasks~\citep{unicorn}. Complementary approaches train models to refine images conditioned on explicit reflections~\citep{reflectionflow}, or aggregate multimodal assessments into rewards for test-time policy optimization~\citep{metattrl}. However, these approaches do not directly distill corrective conditioning into the original-prompt generation policy along its own sampling trajectories. A critique such as ``\textit{restore the missing object}'' describes a desired correction, but does not itself specify how the generator should change its intermediate denoising predictions. 

To bridge this gap, we introduce \textbf{UniEvo-VL}, a novel self-evolving framework that converts visual critique into supervision via \emph{on-policy self-distillation} (OPSD)~\citep{diffusionopd}.  Our mechanism is founded on a widely accepted hypothesis: \emph{verification - examining a solution or an answer - is relatively easier than generation}~\citep{sdpo,zhao2026self}. Toward this goal, we integrate this self-feedback into the vanilla question, producing a revised prompt to help the image generator address the detected failure. Then this new prompt is regarded as privileged information, which only the teacher policy can observe and condition on. This different context makes it possible to induce dense state-wise supervision over the student's sampling trajectories, where the student policy only sees the original prompt. This allows the model to internalize corrective guidance without using corrected images as training targets or scalar rewards for policy optimization.

We instantiate UniEvo-VL with Qwen-Image~\citep{qwenimage}, one well-known open-source flow-based MMM family, on compositional image generation and visual text rendering tasks. The results show that corrective guidance can translate into improved generation from the original prompt alone: direct-generation performance increases from 0.747 to 0.808 on GenEval~\citep{geneval} with Qwen feedback and from 32.97 to 35.53 on GenEval2~\citep{geneval2} Soft-TIFA with the external Luna critic. A paired GenEval evaluation further shows that the evolved generator continues to benefit from an additional reflection pass, suggesting that internalizing corrective experience and using feedback at inference time can be complementary. Together, these findings demonstrate the potential of critique-conditioned self-distillation to retain compositional improvements.

Our contributions are threefold:
\begin{itemize}
    \item \textbf{Critique-conditioned on-policy self-distillation.}
    We introduce UniEvo-VL, which uses critique-derived corrective conditioning to construct teacher predictions along the student's own sampling trajectories. The student learns from the original prompt alone, without corrected-image targets or reward-based policy optimization.

    \item \textbf{Separating learned improvements from inference-time correction.}
    We compare direct and reflection-assisted generation before and after training, distinguishing gains retained in the initial model from the additional benefit of reflection through paired evaluation.

    \item \textbf{Empirical validation and analysis.} We demonstrate improved compositional generation on GenEval and GenEval2, and investigate how critic choice and post-revision verification affect learning across compositional generation and text rendering.
\end{itemize}

\begin{figure}[t]
\centering
\includegraphics[width=\linewidth]{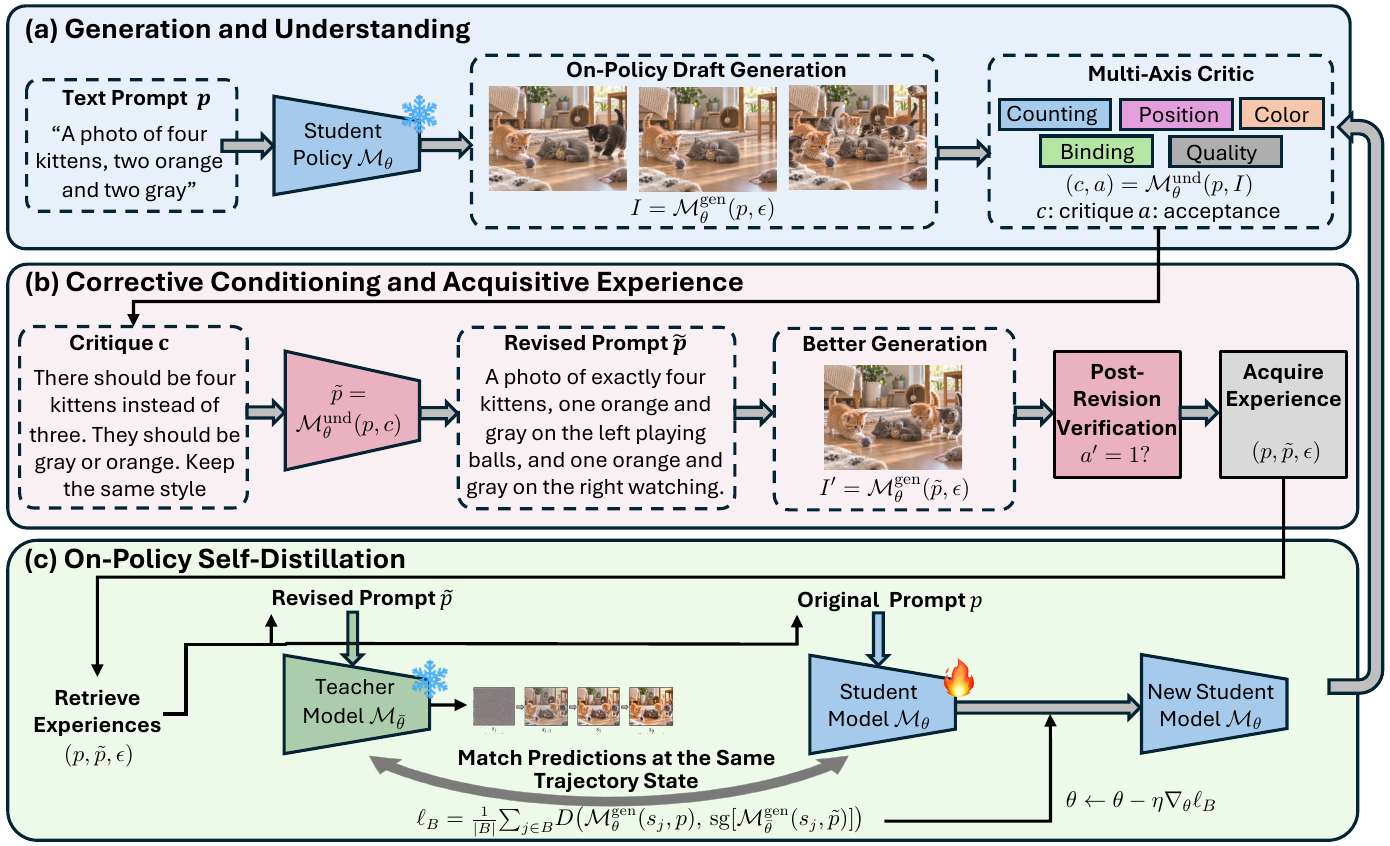}
\caption{\textbf{The \method\ training loop.} A generated draft is inspected along semantic, text, and visual-quality axes. The resulting feedback is synthesized into a revised prompt, which acts as the privileged condition for the EMA teacher. Transition matching updates the student's generator LoRA; the critic is fixed. At deployment, the evolved model uses the original prompt directly.}
\label{fig:framework}
\end{figure}

\section{Method}

\subsection{Self-Evolution Iteration between Generation and Understanding}
\paragraph{Preliminary.} Let $\mathcal{M}_{\boldsymbol{\theta}}$ be a multimodal model (MMM) with two different modes. On the one hand, given a user language prompt $p$, the model can produce an image $I$ from random noise $\boldsymbol{\epsilon}\sim \mathcal{N}(\boldsymbol{0},\boldsymbol{I})$ in the generation mode ($M_{\boldsymbol{\theta}}^{\mathrm{gen}}$). On the other hand, it can also assess any image $I$ against the provided conditioning prompt $p$ in the understanding mode ($M_{\boldsymbol{\theta}}^{\mathrm{und}}$):
\begin{equation}
 I=\mathcal M_{\boldsymbol{\theta}}^{\mathrm{gen}}(p,\boldsymbol{\epsilon}),\qquad
 (c,a)=\mathcal M_{\boldsymbol{\theta}}^{\mathrm{und}}(p,I).
 \label{eq:unified-modes}
\end{equation}
Here, $c$ denotes the discrepancy between the generated image $I$ and its conditioning prompt $p$. $a \in \{0,1\}$ is a binary acceptance decision. For a valid assessment, $a=1$ indicates that the image $I$ highly aligns with $p$ and requires no further revision. Meanwhile, $a=0$ indicates that $I$ is flawed and requires revision. As a remedy, $c$ provides the corresponding corrective feedback. In contrast, empty or unusable critic responses are treated as invalid assessments and will be removed. 

\paragraph{Teacher and student policies.} The joint capability of image generation and understanding paves the road for our \method. It connects these two distinct modes $M_{\boldsymbol{\theta}}^{\mathrm{gen}}$ and $M_{\boldsymbol{\theta}}^{\mathrm{und}}$ through \emph{corrective conditioning}. To achieve this goal, we introduce two roles in the self-evolution loop by varying the conditioning context. The student policy, denoted as $\mathcal{M}_{\boldsymbol{\theta}}$, looks at solely the vanilla prompt $p$, while the reference teacher policy, an EMA version of $\mathcal{M}_{\boldsymbol{\theta}}$ and denoted as $\mathcal{M}_{\bar{\boldsymbol{\theta}}}$, is allowed to access the privileged modification suggestion $c$. 

To begin with, if an image $I$ receives a valid assessment with $a=0$ indicating additional correction, the understanding-mode MMM $M_{\boldsymbol{\theta}}^{\mathrm{und}}$ will consider $c$ as the revision suggestion and update the prompt as $p'$. The teacher policy $\mathcal{M}_{\bar{\boldsymbol{\theta}}}$ uses this privileged prompt $\widetilde p$ to naturally evaluate the student's generation and provide distillation training targets. The student policy $\mathcal{M}_{\boldsymbol{\theta}}$ learns to approach these targets without access to the privileged condition $\widetilde p$.

\subsection{Corrective Conditioning and Experience Acquisition}
\paragraph{Prompt with privileged information.}  For each image $I$ that receives a valid assessment with $a=0$, the understanding-mode MMM $M_{\boldsymbol{\theta}}^{\mathrm{und}}$ synthesizes a privileged prompt $\widetilde{p}$ by incorporating the vanilla prompt $p$ and the corrective feedback $c$:
\begin{equation}
\label{equ:prompt_revision}
    \widetilde{p}=M_{\boldsymbol{\theta}}^{\mathrm{und}}(p,c).
\end{equation}
The revised prompt $\widetilde{p}$ restates the requested scene while explicitly addressing the discrepancies between $I$ and $p$, identified in $c$. For example, an initial prompt $p$ requests two red cups, but $M_{\boldsymbol{\theta}}^{\mathrm{gen}}$ produces a three-cup image. The synthesis process in Equ.~\ref{equ:prompt_revision} is instructed to produce a revised prompt $\widetilde{p}$ that emphasizes exactly two cups while preserving the specified color and scene context.

\paragraph{Prompt filtering with post-hoc evaluation.} After the synthesis of $\widetilde{p}$ by the understanding-mode MMM $M_{\boldsymbol{\theta}}^{\mathrm{und}}$, a naive and straightforward approach is to immediately leverage every $\widetilde{p}$ for subsequent OPSD training. Taking a step further, we introduce a \emph{post-revision verification} stage to determine the feasibility and quality of each $\widetilde{p}$. Explicitly,  we ask the student MMM to generate a new image as $I'=\mathcal{M}_{\boldsymbol{\theta}}^{\mathrm{gen}} (\widetilde{p},\boldsymbol{\epsilon})$ based on the revised prompt $\widetilde{p}$ and the same noise $\boldsymbol{\epsilon}$. Then, this figure $I'$ is fed into the understanding-mode MMM again $\mathcal{M}_{\boldsymbol{\theta}}^{\mathrm{und}}$ for the second-round assessment against $p$. The revised prompt $\widetilde{p}$ will only be selected as a valid training sample if this second-round assessment is valid and accepts $I'$. This extra agreement indicates that $\mathcal{M}_{\boldsymbol{\theta}}^{\mathrm{und}}$ judges $I'$ to satisfy the original prompt $p$ and $I'$ has a better alignment with $p$ than $I$. Notably, the re-generated $I'$ is used only to determine the acceptance of $\widetilde{p}$ and is not regarded as an objective during the subsequent training procedure.

\subsection{On-Policy Self Distillation}
\paragraph{On-policy sampling from the student.} After collecting several valid revised prompts $\widetilde{p}$, we leverage the popular OPSD mechanism to distill the knowledge hidden inside the privileged information~\citep{agarwal2024policy,zhao2026self}. To be specific, at $k$-th iteration with a triple item list $(p,\widetilde{p},\boldsymbol{\epsilon})$, we run a $T$-length diffusion denoising trajectory $\tau= \{s_0, ..., s_T\}$ using the student MMM $\mathcal{M}_{\boldsymbol{\theta}}^{\mathrm{gen}}$ based on the vanilla prompt $p$ and noise $\boldsymbol{\epsilon}$, where $s_j\in \mathbb{R}^d$ denotes the intermediate state in the latent representation space at the $j$-th step in that trajectory. 

Consequently, the student policy $\mathcal{M}_{\boldsymbol{\theta}}^{\mathrm{gen}}$ only observes the prompt statement $p$, matching the inference-time condition. Instead, the teacher policy MMM $\mathcal{M}_{\bar{\boldsymbol{\theta}}}$ conditions on privileged prompt $p'$, which incorporates the modification suggestion $c$ to prevent models from making the same mistakes again. 

\paragraph{Training objective.} We instantiate a \emph{transition divergence objective} that matches the teacher and student denoising distributions at each state $s_j$. Let $\mathcal{M}_{\boldsymbol{\theta}}^{\mathrm{gen}}(s_i,p): \mathbb{R}^d \times \mathcal{P}\rightarrow \mathbb{R}^{d}$ parameterize the velocity field for either flow-based image generation or the denoising score function for diffusion-based algorithms. 

We force the student one-step transition $\mathcal{M}_{\boldsymbol{\theta}}^{\mathrm{gen}}(s_j,p)$ to match the teacher policy's transition target $\mathcal{M}_{\bar{\boldsymbol{\theta}}}^{\mathrm{gen}}(s_j,\widetilde{p})$. The divergence metric elegantly simplifies into a scaled discrepancy strictly between them~\citep{fang2026flow,diffusionopd,yuan2026vision}, written as: 
\begin{equation}
    \mathcal L(\boldsymbol{\theta})  =  \mathbb E_{(p,\widetilde p,\boldsymbol{\epsilon})\sim\mathcal A_k, \tau\sim \mathcal{M}_{\boldsymbol{\theta}}^{\mathrm{gen}}(p,\boldsymbol{\epsilon})),s_j\sim\mathcal \tau} D\!\left( \mathcal{M}_{\boldsymbol{\theta}}^{\mathrm{gen}}(\sg[s_j],p), \sg\!\left[\mathcal{M}_{\bar{\boldsymbol{\theta}}}^{\mathrm{gen}}(\sg[s_j],\widetilde p)\right] \right),
 \label{eq:loss}
\end{equation}
Here, $\mathcal A_k$ denotes the finalized training set, and $D(\cdot,\cdot)$ measures the discrepancy between the student and teacher local predictions, such as KL-divergence or Jensen-Shannon (JS) divergence. $\sg(\cdot)$ stands for stop-gradient backpropagation. Generation, assessment, and synthesis operate without gradients, so gradients only flow to the student policy parameters $\boldsymbol{\theta}$ while the teacher $\mathcal{M}_{\bar{\boldsymbol{\theta}}}^{\mathrm{gen}}$ acts as a fixed full-distribution target. Our flow-based implementation uses squared distance (i.e., mean squared error) between deterministic latent transitions.

\begin{algorithm}[tbp]
\caption{ \textbf{\method.} Critique-conditioned on-policy self-distillation. }
\begin{algorithmic}[1]
\scriptsize
\Statex \textbf{Input:} model $\mathcal M_{\theta}$, prompts $\mathcal P$,
verification $\nu$, update rule $\mathcal R$ 
\vspace{3pt} \hrule \vspace{3pt} 

\State $\bar{\theta}\gets\theta$
\For{$k=0,1,\ldots$}
    \State $\mathcal A_k\gets\varnothing$

    \Statex \textit{\textbf{Corrective experience acquisition}}
    \For{$p\in\mathcal P$}
        \State $\epsilon\sim\mathcal N(0,I)$
        \State $I\gets\mathcal M_{\theta}^{\mathrm{gen}}(p,\epsilon)$
        \State $(c,a)\gets\mathcal M_{\theta}^{\mathrm{und}}(p,I)$

        \If{$c$ valid and $a=0$}
            \State $\tilde p\gets
            \mathcal M_{\theta}^{\mathrm{und}}(p,c)$

            \If{$\nu=1$}
                \State $I'\gets
                \mathcal M_{\theta}^{\mathrm{gen}}(\tilde p,\epsilon)$
                \State $(c',a')\gets
                \mathcal M_{\theta}^{\mathrm{und}}( p,I')$
                \If{$c'$ valid and $a'=1$}
                    \State $\mathcal A_k\gets
                    \mathcal A_k\cup\{(p,\tilde p,\epsilon)\}$
                \EndIf
            \Else
                \State $\mathcal A_k\gets
                \mathcal A_k\cup\{(p,\tilde p,\epsilon)\}$
            \EndIf
        \EndIf
    \EndFor

    \Statex \textit{\textbf{On-policy self-distillation}}
    \For{$(p,\tilde p,\epsilon)\in\mathcal A_k$}
        \State $\tau\gets
        \sg[\mathcal M_{\theta}^{\mathrm{gen}}(p,\epsilon)]$

        \For{minibatch $B\subset\tau$}
            \State $\ell_B\gets\frac{1}{|B|}
            \sum_{j\in B} D\!\left(
            \mathcal M_{\theta}^{\mathrm{gen}}(s_j,p),
            \sg[\mathcal M_{\bar\theta}^{\mathrm{gen}}
            (s_j,\tilde p)]\right)$
            \State $\theta\gets\theta-\eta\nabla_\theta\ell_B$
            \State $\bar\theta\gets
            \mathcal R(\bar\theta,\theta)$
        \EndFor
    \EndFor
\EndFor
\State \Return $\mathcal M_\theta$
\end{algorithmic}
\label{alg:unievo}
\end{algorithm}

\paragraph{Discussion.} On-policy sampling places supervision at states $s_j\in \tau$ visited by the current student. The privileged prompt $\widetilde p$ provides the teacher policy with corrective information, which in turn guides the student toward denoising paths that lead to the correct answer. Equ.~\ref{eq:loss} thereby transfers this privileged correction into the student parameters $\boldsymbol{\theta}$, enabling direct inference using the original prompt $p$ without requiring any sort of revision or reflection $c$.

Alg.~\ref{alg:unievo} summarizes the procedure. The student MMM $\mathcal{M}_{\boldsymbol{\theta}}^{\mathrm{gen}}$ carries its updated parameters forward to subsequent requests, allowing corrective experience to accumulate across the request stream during test time. The teacher's parameters $\bar{\boldsymbol{\theta}}$ are held fixed for each minibatch iteration and refreshed according to a reference update rule $\mathcal R$. The update rule $\mathcal R$ and trajectory-step selection principles are specified in Appendix~\ref{app:config}.

\section{Experiments}
\label{sec:experiments}
\subsection{Experimental Setup}

\paragraph{Tasks and implementation.} Qwen-Image family incorporates Qwen-VL family as conditioning encoders. We therefore selected Qwen-Image-2512~\citep{qwenimage} and a fixed Qwen-VL~\citep{bai2025qwen3} feedback pipeline to examine an architecturally motivated generation–understanding setting. To test generalization to a stronger generator backbone, we additionally evaluate UniEvo-VL with Qwen-Image-2.1 on GenEval2 across three training seeds (42, 43, and 44). During training, only generator LoRA~\citep{hu2021lora} parameters are updated, separately for GenEval~\citep{geneval}, GenEval2~\citep{geneval2}, and OCR text rendering. We evaluate prompt filtering with and without post-revision verification. Section~\ref{sec:critic} further investigates whether stronger external critique can yield greater improvements using GPT-5.6-Luna. Appendices~\ref{app:config}, and~\ref{app:datasets} provide implementation and dataset details.   Training compute and acquisition overhead are discussed in Appendix~\ref{app:training-compute}.

\paragraph{Evaluation.}
We report direct generation and inference with one critique
opportunity separately, assessing outputs against the original
prompts and pairing prompts and seeds within each experiment.
Metrics comprise native GenEval and OCR scores (0--1),
GenEval2 Soft-TIFA GM (0--100), Gemini atomic accuracy (\%)
for compositional tasks, and holistic Gemini task and HumanPref
scores (0--10). HumanPref is an automated visual-quality proxy.
Evaluation scores provide no training reward.
Appendix~\ref{app:eval} specifies evaluation cohorts, metric
definitions, and failure handling; checkpoint coverage is
provided in the Supplementary Data.

\begin{figure}[t]
\centering
\includegraphics[width=\linewidth]{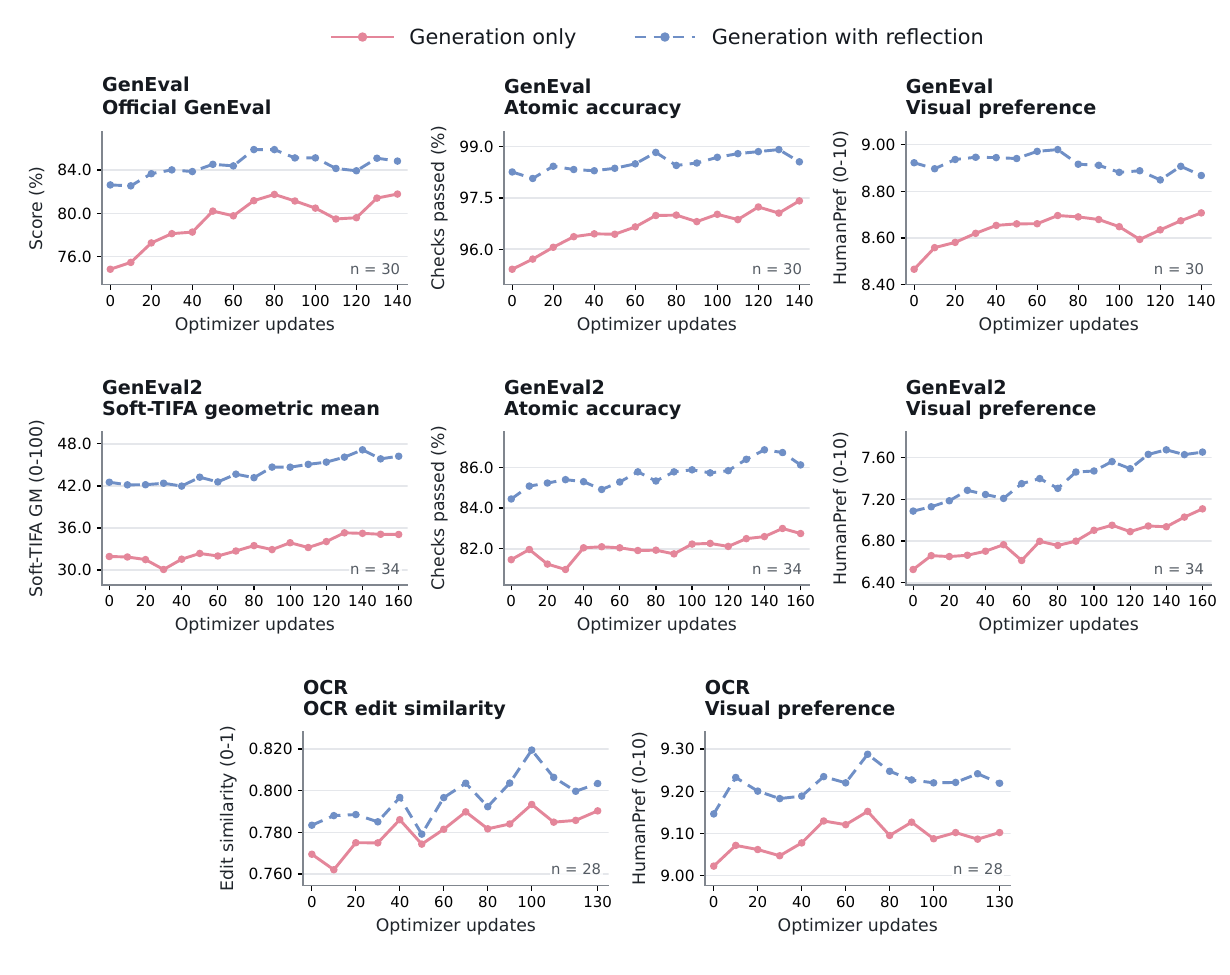}
\vspace{-1.5em}
\caption{\textbf{UniEvo-VL with revision verification across training updates.}
Solid curves show direct generation; dashed curves allow one inference-time critique. Lines connect evaluated checkpoints across runs; $n$ counts plotted points.  }
\label{fig:overview}
\end{figure}

\subsection{Unified Direct-Generation Results}
\begin{table*}[t]
\centering
\small
\caption{\textbf{Performance of UniEvo-VL.}
Parentheses identify the critic; ``+ verification'' denotes revision-prompt filtering. Base is averaged over different random seeds of the identical pretrained model. \textbf{Bold} and \underline{underlined} values indicate the best and second-best for each dataset and metric.}
\label{tab:main-results}

\setlength{\tabcolsep}{4.2pt}
\renewcommand{\arraystretch}{1.08}

\resizebox{\textwidth}{!}{
\begin{tabular}{l ccc ccc ccc}
\toprule
& \multicolumn{3}{c}{\textbf{GenEval}}
& \multicolumn{3}{c}{\textbf{GenEval2}}
& \multicolumn{3}{c}{\textbf{OCR}} \\
\cmidrule(lr){2-4}
\cmidrule(lr){5-7}
\cmidrule(lr){8-10}

Method
& Native & Atomic (\%) & HumanPref
& Native & Atomic (\%) & HumanPref
& Native & Atomic (\%) & HumanPref \\
\midrule

Base
& 0.747 & 95.30 & 8.48
& 32.58 & 81.69 & 6.65
& 0.771 & -- & \underline{9.07} \\

UniEvo-VL
& 0.808 & 96.90 & \underline{8.79}
& 32.37 & 82.24 & \underline{6.91}
& 0.761 & -- & 9.06 \\

UniEvo-VL (GPT5.6-Luna)
& \textbf{0.882} & \textbf{98.76} & \textbf{8.97}
& \textbf{35.53} & \textbf{83.18} & 6.86
& \underline{0.775} & -- & 9.03 \\

UniEvo-VL (verification)
& \underline{0.818} & \underline{97.41} & 8.71
& \underline{35.07} & \underline{82.75} & \textbf{7.11}
& \textbf{0.790} & -- & \textbf{9.10} \\

\bottomrule
\end{tabular}
}
\end{table*}

Table~\ref{tab:main-results} compares image generation performance after OPSD. The Qwen configuration with revision verification exceeds the reported Base on every listed metric across all three tasks, although different configurations lead on individual metrics.

\paragraph{Compositional generation.} On GenEval, both Qwen configurations score above Base on semantic correctness and automated visual quality. The external-critic experiment with GPT-5.6-Luna achieves the highest native score, 0.882, as well as the highest atomic accuracy and HumanPref rating. Among the Qwen configurations, verified Qwen scores higher on native score and atomic accuracy, whereas unverified Qwen receives a higher HumanPref rating. GenEval2 shows a broadly similar pattern: the external-critic experiment with GPT-5.6-Luna again achieves the highest native score, 35.53, and atomic accuracy, while verified Qwen achieves the highest HumanPref rating. Unverified Qwen scores above Base on atomic accuracy and HumanPref but slightly below it on native Soft-TIFA. Thus, the evaluations distinguish improvements in individual semantic checks, benchmark-level correctness, and visual quality.

\paragraph{Text rendering.}
The verified Qwen configuration also leads on OCR, reaching a native
score of 0.790 and the highest HumanPref rating. The other
configurations show mixed outcomes: unverified Qwen falls below
Base on both measures, while Luna scores slightly higher on native
text fidelity but lower on HumanPref. The favorable OCR result
therefore does not extend to every feedback configuration.

\begin{figure}[t]
\centering
\includegraphics[width=\linewidth]{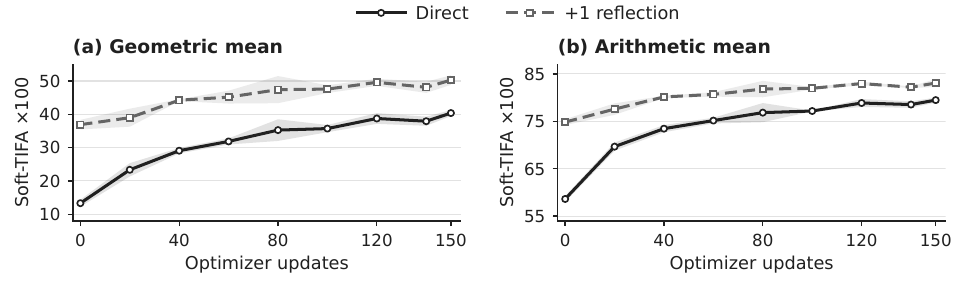}
\caption{Qwen-Image-2.1 learning curves on official GenEval2. Lines show means and bands show $\pm1$ sample SD across seeds 42, 43, and 44; each endpoint uses 800 images and 6,012 checks. Direct generation uses the original prompt, and +1 reflection adds one correction pass. Full checkpoint values appear in Table~\ref{tab:qwen21-full}.}
\label{fig:qwen21-curves}
\end{figure}

\begingroup
\paragraph{Generalization to Qwen-Image-2.1.}
Corrective adaptation remains effective with the stronger Qwen-Image-2.1 backbone. On GenEval2, direct-generation Soft-TIFA GM improves from $13.30\pm1.04$ to $40.37\pm0.89$ after 150 updates across seeds 42, 43, and 44 (mean $\pm$ sample SD), with a positive gain in every seed. One additional reflection reaches $50.27\pm1.43$ GM. Figure~\ref{fig:qwen21-curves} shows both learning curves. Appendix~\ref{app:qwen21-controls} reports all checkpoints, arithmetic-mean scores, configurations, and prompt-only and verification controls.
\endgroup

\subsection{Improvement and Preservation Across Prompt Difficulty}
\label{sec:difficulty}

We examine where training improves generation and how it affects
initially successful outputs. Using historical 500-prompt subsets
for each task, we define Hard and Easy groups by baseline holistic
task scores of 0--4 and 5--10, respectively. Group membership remains
fixed after training; evaluation details are provided in
Appendix~\ref{app:historical-difficulty}.

\begin{figure}[t]
\centering
\includegraphics[width=\linewidth]{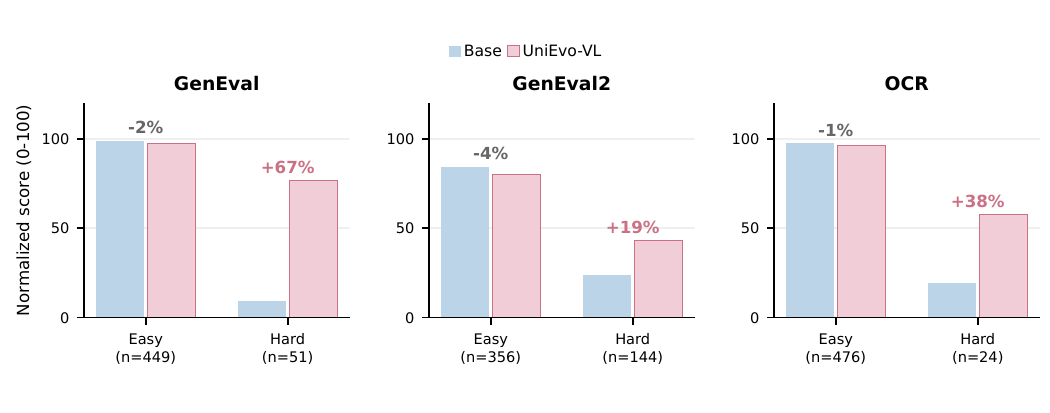}
\vspace{-3em} 
\caption{Improvement and preservation on historical 500-prompt subsets, where scale is normalized to 0--100. The percentage-point changes are subset-specific, not full-test-set gains.}
\label{fig:historical-difficulty}
\end{figure}

Figure~\ref{fig:historical-difficulty} shows that gains concentrate
on prompts the base model initially struggles with across all
three tasks. Average performance on Easy prompts remains high,
but decreases by 1--4 points on the normalized 0--100 scale.
Training therefore combines substantial recovery on initially
low-scoring prompts with small regressions on higher-scoring ones.

This concentration of gains is partly expected from our acquisition
mechanism. Drafts judged satisfactory by the critic are skipped,
while valid corrective pairs from drafts requiring revision supply
the training signal. The resulting selection emphasizes current
generation failures, creating more opportunities to learn corrective
behavior than to reinforce already satisfactory outputs.
The observed difficulty pattern is consistent with this mechanism,
although the evaluation groups are defined by baseline task scores
rather than acquisition decisions. Their different improvement
headroom also prevents attributing the contrast to filtering alone.

\subsection{Understanding Self-Evolution}
\subsubsection{Does Training Transfer the Benefit of Reflection?}
\label{sec:gap}

Table~\ref{tab:gap} compares the base model and UniEvo-VL with direct generation
or one reflection opportunity. The evaluated UniEvo-VL checkpoints use post-revision
verification during training; all outputs are scored against the original
request.

\begin{table}[t]
\centering
\caption{\textbf{Direct generation and reflection before and after training.}
(a) Generation scores, where bold marks the highest score in each row. (b) Training gain is the UniEvo-VL score minus the base score under direct
generation, while reflection gain is reflected minus direct within each model.
Gains use unrounded scores. Native scores use 0--1 for GenEval/OCR and
0--100 for GenEval2. Atomic uses percent (gains in percentage points),
and HumanPref uses 0--10.}
\label{tab:gap}
\label{tab:verified-reflection}
\begingroup
\small
\setlength{\tabcolsep}{5pt}
\renewcommand{\arraystretch}{1.12}
\begin{tabular*}{\linewidth}{@{\extracolsep{\fill}}llrrrr@{}}
\toprule
\textbf{(a)} & & \multicolumn{2}{c}{Base model} & \multicolumn{2}{c}{UniEvo-VL}\\
\cmidrule(lr){3-4}\cmidrule(l){5-6}
Dataset & Metric & Direct & + Reflection & Direct & + Reflection\\
\midrule
\textbf{GenEval} & Native & $0.748$ & $0.826$ & $0.818$ & $\mathbf{0.848}$\\
 & Atomic & $95.41$ & $98.26$ & $97.41$ & $\mathbf{98.56}$\\
 & HumanPref & $8.47$ & $\mathbf{8.92}$ & $8.71$ & $8.87$\\
\midrule
\textbf{GenEval2} & Native & $31.92$ & $42.52$ & $35.07$ & $\mathbf{46.23}$\\
 & Atomic & $81.47$ & $84.45$ & $82.75$ & $\mathbf{86.12}$\\
 & HumanPref & $6.53$ & $7.09$ & $7.11$ & $\mathbf{7.65}$\\
\midrule
\textbf{OCR} & Native & $0.769$ & $0.783$ & $0.790$ & $\mathbf{0.803}$\\
 & HumanPref & $9.02$ & $9.15$ & $9.10$ & $\mathbf{9.22}$\\
\bottomrule
\end{tabular*}
\par\vspace{0.7em}
\begin{tabular*}{\linewidth}{@{\extracolsep{\fill}}llrrr@{}}
\toprule
\textbf{(b)} & & Training gain & \multicolumn{2}{c}{Reflection gain}\\
\cmidrule(lr){3-3}\cmidrule(l){4-5}
Dataset & Metric & Direct generation & Base model & UniEvo-VL\\
\midrule
\textbf{GenEval} & Native & $+0.069$ & $+0.078$ & $+0.030$\\
 & Atomic & $+2.00$ & $+2.85$ & $+1.14$\\
 & HumanPref & $+0.24$ & $+0.46$ & $+0.16$\\
\midrule
\textbf{GenEval2} & Native & $+3.15$ & $+10.60$ & $+11.17$\\
 & Atomic & $+1.28$ & $+2.98$ & $+3.37$\\
 & HumanPref & $+0.58$ & $+0.56$ & $+0.54$\\
\midrule
\textbf{OCR} & Native & $+0.021$ & $+0.014$ & $+0.013$\\
 & HumanPref & $+0.08$ & $+0.12$ & $+0.12$\\
\bottomrule
\end{tabular*}
\endgroup
\end{table}

\paragraph{OPSD improves inference-time direct generation.}
Table~\ref{tab:gap} shows that training improves direct generation on
every reported metric across all three benchmarks. On native scores,
direct generation with UniEvo-VL approaches the reflected base model on GenEval
and exceeds it on OCR, while remaining below it on GenEval2. Training
therefore makes improvements available from the original prompt alone,
although it does not consistently recover the performance achieved by
applying reflection to the base model.

\paragraph{Reflection remains useful after training.}
Reflection further improves the evolved model on every reported metric.
On GenEval, its native gain decreases from $0.078$ to $0.030$, with
similar reductions in atomic accuracy and HumanPref gains. This pattern
is consistent with partly overlapping benefits from training and
reflection. By contrast, reflection gains remain substantial on GenEval2
and change little on OCR. Improved direct generation therefore need not
be accompanied by a smaller reflection gap: training can improve the
generator while leaving considerable benefit from additional feedback.

\paragraph{Capability emerges beyond inference-time reflection.}
Importantly, the retained training benefit cannot be explained solely by
applying prompt revision at inference time. Under the same one-reflection
protocol, UniEvo-VL improves the native score over the reflected base model
from $0.826$ to $0.848$ on GenEval, from $42.52$ to $46.23$ on GenEval2,
and from $0.783$ to $0.803$ on OCR. Thus, corrective training provides
improvements that persist even when both models are given an additional
opportunity for reflection, indicating acquired capability beyond
inference-only prompt revision. The combined setting achieves the highest
native scores on all three benchmarks, although GenEval HumanPref remains
slightly higher for the reflected base model.

\subsubsection{How Does Critic Capacity Affect Self-Evolution?}
\label{sec:critic}
We examine how feedback choice affects compositional generation on
GenEval and GenEval2, using acquisition without post-revision verification
and generation only at evaluation. Table~\ref{tab:critic} reports
category scores and, where matched baselines are available, the gains
retained after self-evolution.

\begin{table}[t]
\centering
\caption{\textbf{Compositional generation under different critics.}
All entries are Gemini compositional-fidelity ratings (0--10).
(a) GenEval uses 553 prompts; Base denotes the Qwen-VL experiment's
baseline. (b) GenEval2 uses 800 prompts; $\Delta$ is Evolved minus Base.
Bold indicates the highest score in each column of (a), and the
higher Evolved score and larger $\Delta$ within each row of (b),
including ties.}
\label{tab:critic}
\begingroup
\small
\setlength{\tabcolsep}{3pt}
\renewcommand{\arraystretch}{1.08}

\textbf{(a) GenEval}\par\smallskip
\begin{tabular*}{\linewidth}{@{\extracolsep{\fill}}lrrrrrrr@{}}
\toprule
Critic / model & Single & Two & Count & Color & Position & Attr. & Overall\\
\midrule
Base
& 9.93 & 9.68 & 7.86 & 9.71 & 6.85 & 9.75 & 8.96\\
Qwen-VL
& 9.90 & 9.88 & 9.29 & 9.66 & 8.79 & \textbf{9.94} & 9.57\\
GPT-5.6-Luna$^\dagger$
& \textbf{9.98} & \textbf{9.92} & \textbf{9.58}
& \textbf{9.91} & \textbf{9.88} & \textbf{9.94}
& \textbf{9.87}\\
\bottomrule
\end{tabular*}

\par\medskip
\textbf{(b) GenEval2}\par\smallskip
\begin{tabular*}{\linewidth}{@{\extracolsep{\fill}}lrrrrrr@{}}
\toprule
& \multicolumn{3}{c}{Qwen-VL}
& \multicolumn{3}{c}{GPT-5.6-Luna}\\
\cmidrule(lr){2-4}\cmidrule(l){5-7}
Category & Base & Evolved & $\Delta$
         & Base & Evolved & $\Delta$\\
\midrule
Two objects
& 7.62 & 7.90 & $+0.28$
& 7.00 & \textbf{8.21} & $\mathbf{+1.21}$\\
Color
& 6.27 & 6.64 & $+0.37$
& 6.54 & \textbf{7.56} & $\mathbf{+1.02}$\\
Position
& 5.68 & 5.74 & $+0.06$
& 5.78 & \textbf{5.92} & $\mathbf{+0.14}$\\
Attribute
& 7.13 & 7.17 & $+0.04$
& 7.22 & \textbf{7.28} & $\mathbf{+0.06}$\\
\addlinespace[2pt]
Overall
& 6.14 & 6.23 & $+0.09$
& 6.23 & \textbf{6.45} & $\mathbf{+0.22}$\\
\bottomrule
\end{tabular*}
\endgroup
\end{table}

\paragraph{GenEval differences are largest in spatial composition.}
Qwen feedback improves position scores from 6.85 to 8.79 and counting
from 7.86 to 9.29. The historical Luna result reaches 9.88 and 9.58,
respectively, with position showing the largest gap between evolved
configurations. Several other categories are already near saturation.

\paragraph{GenEval2 reveals category-dependent learning gains.}
Luna yields larger gains in two-object composition and color:
approximately $+1.21$ and $+1.02$, versus $+0.28$ and $+0.37$ for Qwen.
For two objects, Luna starts lower (7.00 versus 7.62) but finishes higher
(8.21 versus 7.90). Position and attribute scores improve only modestly
under either configuration. Native Soft-TIFA GM (0--100) corroborates the
overall pattern, increasing from 32.97 to 35.53 with Luna, versus 32.86
to 32.37 with Qwen. The observed benefit of external feedback is skill-dependent.
Luna achieves a higher overall GenEval endpoint and larger two-object
and color gains on GenEval2, while GenEval2 position and attribute
scores show limited gains. Category-level analysis is therefore essential
for identifying the capabilities a vision language model need to possess for steady improvement through
self-evolution.

\section{Related Work}
\label{sec:related}

\paragraph{Understanding and reflection for image generation.}
Visual understanding helps models assess and improve their generated images.
\citet{internalgap} use the understanding branch to score images and construct supervised fine-tuning and preference data.
SRUM~\citep{srum} combines image-level and object-level self-rewards for reward-weighted training, while UniCorn~\citep{unicorn} assigns proposer, solver, and judge roles to a unified model to generate training interactions.
ReflectionFlow~\citep{reflectionflow} learns from flawed-image, reflection, and improved-image triplets for iterative inference-time refinement.
UniReason~\citep{unireason} combines world-knowledge reasoning before generation with visual refinement afterward through two-stage supervised fine-tuning on curated examples.
\method\ transfers corrective feedback into direct generation from the original request and measures the additional benefit of reflection after training.

\paragraph{Feedback-driven optimization.}
Meta-TTRL~\citep{metattrl} aggregates model-generated rubric assessments into rewards for test-time policy optimization.
Flow-GRPO~\citep{flowgrpo} and DiffusionNFT~\citep{diffusionnft} improve diffusion or flow generators through reward-based updates.
DiffusionOPSD~\citep{diffusionopsd} converts image-level reward gradients into detached clean-output targets.
\method\ also incorporates feedback into model parameters, but derives targets from teacher predictions conditioned on corrective text, requiring neither gradients through the critic nor a differentiable image reward.

\paragraph{On-policy distillation with privileged conditioning.}
OPD~\citep{agarwal2024policy} trains a student on its own trajectories using teacher predictions.
SDPO~\citep{sdpo} distills feedback-conditioned next-token predictions into a language policy.
For diffusion models, DiffusionOPD~\citep{diffusionopd} matches teacher--student transitions on student trajectories to consolidate independently trained, task-specific teachers.
D-OPSD~\citep{dopsd} conditions a diffusion self-teacher on the target image and text for supervised learning on student rollouts while preserving few-step generation.
Dreaming in Flow~\citep{ggf} combines image-grounded and repair-enhanced flow targets with dream replay to jointly improve understanding and generation.
\method\ derives privileged conditioning from critiques of its own generated images: the teacher receives a revised prompt, while the student receives the original request.
Their denoising predictions are matched along student trajectories without paired target images or an independently task-trained generator teacher.
Only the generator is updated, with a fixed critic, to improve subsequent direct generation.

\section{Conclusion}
We introduced UniEvo-VL, an approach to multimodal self-evolution that
uses on-policy self-distillation to turn the corrective content of visual
critique into supervision for generation. Paired evaluations on GenEval,
GenEval2, and OCR show that training improves native scores both with
and without reflection, with the combination of training and reflection
performing best on these metrics. Gains vary across training
configurations, and regressions on initially easier prompts reveal limits
to preservation. These findings suggest a route to self-evolution in
which feedback-guided generation supplies supervision and improves
alongside the model's unassisted capability.



\subsection*{AI use statement}
We used generative AI tools to assist with manuscript drafting and
editing, literature search and organization, and code and figure
preparation. The authors reviewed and revised AI-assisted material and
checked adopted code and factual claims before inclusion.
Language and multimodal models also serve as critics,
prompt-synthesis models, and automated evaluators in our experiments.
These experimental uses, including the model configurations,
prompts, and evaluation procedures, are documented in the main
text and appendices. The authors take full responsibility for the
final manuscript, implementation, reported results, and scientific
conclusions.

\section*{Reproducibility Statement}
The training objective and algorithm are described in
Section~2 and Appendix~A. Appendix~A also specifies the model
configurations, training hyperparameters, sampling settings, random
seeds, and computational requirements. Dataset sources and construction
are described in Appendix~C, while Appendix~D documents the evaluation
cohorts, inference settings, metric definitions, and handling of failed
evaluation requests. Appendix~G provides the critic and prompt-synthesis
templates used in the Qwen configuration with revision verification.

\section*{Acknowledgment}

We acknowledge computational resources provided by Stanford's
Marlowe GPU computing instrument~\cite{marlowe2025}, supported
by Stanford HAI and Stanford Research Computing.

\bibliography{references}
\bibliographystyle{iclr2027_conference}

\appendix
\newpage 

\section{Implementation and Configuration}
\label{app:config}

\paragraph{Choice of generation and understanding models}

We select Qwen-Image-2512 and Qwen-VL to study critique-guided improvement in an architecturally motivated generation–understanding setting. Qwen-Image-2512 already incorporates a Qwen2.5-VL conditioning encoder, making the Qwen-VL family a natural choice for investigating how visual understanding can support improvements in image generation. Our objective is to convert explicit corrective feedback into updates that improve generation from the original prompt, rather than use feedback only for inference-time revision. Some experiments without verification ablate using GPT-5.6-Luna feedback. Training updates only image-generation LoRA parameters. The base generation weights, text encoder, VAE, and feedback-producing components remain fixed. 

\subsection{Flow-based generation objective}
\label{app:flow-objective}

We instantiate the local prediction in Section~2.3 with a flow velocity.
Let $s_j\in\mathbb R^d$ be the noisy latent at step $j$ of the student
trajectory, let $\sigma_j$ be its noise level, and write
$\Delta\sigma_j=\sigma_{j+1}-\sigma_j$.
The notation $\mathcal M_{\boldsymbol{\theta}}^{\mathrm{gen}}(p,\epsilon)$ denotes complete
image generation, whereas $\mathcal M_{\boldsymbol{\theta}}^{\mathrm{gen}}(s_j,p)$ denotes
a local velocity prediction, with $\sigma_j$ implicit in the step index.
Specifically, the implementation uses classifier-free guidance:
\begin{equation}
\begin{aligned}
v_{\boldsymbol{\theta}}^{(g)}(z,\sigma,p)
&=v_{\boldsymbol{\theta}}(z,\sigma,\varnothing)
 +g\bigl[v_{\boldsymbol{\theta}}(z,\sigma,p)-v_{\boldsymbol{\theta}}(z,\sigma,\varnothing)\bigr],\\
\mathcal M_{\boldsymbol{\theta}}^{\mathrm{gen}}(s_j,p)
&=v_{\boldsymbol{\theta}}^{(g)}(s_j,\sigma_j,p),\qquad g=4,
\end{aligned}
\label{eq:flow-cfg}
\end{equation}
where $v_{\boldsymbol{\theta}}$ is the unguided velocity predictor and $\varnothing$
denotes empty text conditioning. The same guidance rule is used for the
student and reference predictions; these training predictions receive
no additional norm rescaling.
The predicted next latent is
\begin{equation}
\mu_{\theta,j}(z,p)
=z+\Delta\sigma_j\,v_{\boldsymbol{\theta}}^{(g)}(z,\sigma_j,p).
\label{eq:flow-transition}
\end{equation}

For each acquired triple $(p,\widetilde p,\epsilon)\in\mathcal A_k$,
we collect the detached trajectory
$\tau=\operatorname{sg}[\operatorname{Rollout}_{\boldsymbol{\theta}}(p,\epsilon)]$
under the original prompt. Student and reference predictions use the
same detached latent and noise level, with conditioning $p$ and
$\widetilde p$, respectively. The per-transition loss is
\begin{equation}
\begin{aligned}
\ell_j^{\mathrm{flow}}(\boldsymbol{\theta})
&=\frac{1}{2d}\left\|
\mu_{\theta,j}(\operatorname{sg}[s_j],p)
-\operatorname{sg}\!\left[
\mu_{\bar\theta,j}(\operatorname{sg}[s_j],\widetilde p)
\right]\right\|_2^2\\
&=\frac{(\Delta\sigma_j)^2}{2d}\left\|
\mathcal M_{\boldsymbol{\theta}}^{\mathrm{gen}}(\operatorname{sg}[s_j],p)
-\operatorname{sg}\!\left[
\mathcal M_{\bar\theta}^{\mathrm{gen}}(\operatorname{sg}[s_j],\widetilde p)
\right]\right\|_2^2.
\end{aligned}
\label{eq:flow-loss}
\end{equation}
The second equality follows because the shared latent cancels.
Thus, the flow instantiation of the main-method discrepancy is
$\mathcal D_j(u,w)=(\Delta\sigma_j)^2\|u-w\|_2^2/(2d)$.
This is deterministic transition matching, equivalently velocity
matching weighted by the squared sampler interval; it is not an exact
stochastic-policy KL or JS divergence.

Let $\mathcal J\subseteq\{0,\ldots,T-1\}$ denote the selected transition
indices. The objective is
\begin{equation}
\mathcal L_{\mathrm{flow}}(\boldsymbol{\theta})
=\mathbb E_{(p,\widetilde p,\epsilon)\sim\mathcal A_k}
\left[\frac{1}{|\mathcal J|}
\sum_{j\in\mathcal J}\ell_j^{\mathrm{flow}}(\boldsymbol{\theta})\right].
\label{eq:flow-objective}
\end{equation}
For the reported 20-step, noisy-fraction-$0.3$ recipe,
$\mathcal J=\{0,\ldots,5\}$ selects the six highest-noise transitions
of the configured scheduler grid.
Gradients pass only through the student's local predictions. Draft
generation, rollout collection, assessment, synthesis, and reference
targets are detached. Gradients are accumulated over the selected
transitions and configured example batches; the EMA reference is updated
after actual optimizer steps, as described in Appendix~A.2.
Draft generation and the corresponding training rollout reuse the
acquired seed. The regenerated image $I'$ is used only for verification
in the verified OPSD variant and is not a training target for this loss.

The general framework requires architecture-compatible local predictions.
An autoregressive extension would need a separately defined loss on
next-token distributions at a shared prefix and vocabulary; our
experiments validate only the flow instantiation above.

\subsection{Acquisition and reference parameters}
\label{app:acquisition}

An initially accepted draft supplies no corrective pair and is skipped.
For a draft that receives valid corrective feedback, the synthesis module
constructs a revised prompt $\widetilde p$ from the original prompt $p$ and
the critique. Acquisition without post-revision verification admits the
resulting triple $(p,\widetilde p,\epsilon)$ directly.

Acquisition with post-revision verification permits one revision. We
regenerate
\[
I' = \mathcal{M}^{\mathrm{gen}}_{\theta}(\widetilde p,\epsilon)
\]
using the same generation seed as the initial draft, and assess $I'$ against
the \emph{original} prompt $p$. The triple $(p,\widetilde p,\epsilon)$ is
admitted for distillation only when this second assessment is valid and
accepts $I'$ under $p$. Malformed feedback, invalid revisions, and revisions
whose regenerated images remain unaccepted under the original request are
discarded.

The regenerated image $I'$ is used only for acquisition filtering and never
serves as a distillation target. During OPSD, $\widetilde p$ instead provides
the privileged condition to the reference generator, while the student
remains conditioned on the original prompt $p$. 

The reference generation adapter is updated after optimizer steps using EMA as \(\bar{\theta}
\leftarrow
\beta \bar{\theta} + (1-\beta)\theta,
\) with $\beta=0.999$ in the main experiments. The base generator weights remain
shared and fixed, and all reference predictions are computed without
gradients. Appendix~\ref{app:teacher} reports the EMA-decay
sweep.

\subsection{Two Training Recipes}
Tables~\ref{tab:config} and~\ref{tab:confirmed-config} specify the training recipes without and with post-revision verification, respectively. The recipes differ in feedback backend, acquisition seeds, learning rate, and gradient accumulation. Table~\ref{tab:config} includes the unverified Qwen counterparts on GenEval2 and OCR alongside the Luna configurations. The GenEval configuration shown is the recorded recipe; checkpoint metadata do not independently encode the full launch configuration.

\begin{table}[htbp]
\centering\small
\caption{Training configuration for acquisition without post-revision verification.}
\label{tab:config}
\setlength{\tabcolsep}{3pt}\begin{tabular}{@{}lccccc@{}}
\toprule
Setting & GenEval & \multicolumn{2}{c}{GenEval2} & \multicolumn{2}{c}{OCR}\\
\midrule
Feedback backend & Qwen3-VL & Qwen3-VL & GPT-5.6-Luna & Qwen3-VL & GPT-5.6-Luna\\
Critic parameters & 8B & 8B & - & 8B &  - \\
Generator & \multicolumn{5}{c}{Qwen-Image-2512}\\
Resolution / steps / CFG & \multicolumn{5}{c}{$1024^2$ / 20 / 4.0}\\
LoRA rank / alpha & \multicolumn{5}{c}{16 / 16}\\
Learning rate & $3\times10^{-4}$ & $1\times10^{-4}$ & $1\times10^{-4}$ & $3\times10^{-4}$ & $3\times10^{-4}$\\
Weight decay & \multicolumn{5}{c}{$10^{-4}$}\\
Local batch / accumulation & 1 / 1 & 1 / 2 & 1 / 1 & 1 / 2 & 1 / 1\\
Seeds per acquired prompt & 1 & 8 & 8 & 1 & 1\\
Correction opportunities & \multicolumn{5}{c}{1}\\
Inner epochs per batch & \multicolumn{5}{c}{1}\\
Noisy timestep fraction & \multicolumn{5}{c}{0.3}\\
Teacher EMA decay / run seed & \multicolumn{5}{c}{0.999 / 0}\\
Post-revision verification & \multicolumn{5}{c}{No}\\
\bottomrule
\end{tabular}
\end{table}

\begin{table}[htbp]
\centering\small
\caption{Training configuration for acquisition with post-revision verification on all three tasks. Gradient accumulation operates over transition-loss evaluations.}
\label{tab:confirmed-config}
\begin{tabular}{@{}ll@{}}
\toprule
Setting & Value\\
\midrule
Generator & Qwen-Image-2512\\
Feedback backend & Qwen3-VL 8B, short-panel critique\\
Resolution / steps / CFG & $1024^2$ / 20 / 4.0\\
LoRA rank / alpha & 16 / 16\\
Optimizer / learning rate & AdamW / $3\times10^{-4}$\\
Adam coefficients / epsilon & $(0.9,0.999)$ / $10^{-8}$\\
Weight decay / gradient clipping & $10^{-4}$ / 1.0\\
Local batch / accumulation & 1 / 2\\
Training hardware & Four H200 GPUs\\
Acquisition seeds per prompt & 1\\
Transition selection & Noisiest 30\% (six of 20)\\
Inner epochs per batch & 1\\
EMA decay / run seed & 0.999 / 42\\
Post-revision verification & Required\\
\bottomrule
\end{tabular}
\end{table}

The configured feedback pipeline for acquisition with verification uses Qwen3-VL-8B-Thinking for critique and Qwen3-VL-8B-Instruct for synthesis. Gemini-2.5-Flash is used only for evaluation.

\paragraph{Inference conditions.}
Direct and one-critique evaluations both load the saved student adapter. One-critique inference retains an accepted draft and otherwise revises and regenerates it. Every benchmark prompt is scored against its original request, without filtering the evaluation population by the training acceptance rule. Equal critique budgets can incur different computation because accepted drafts require no regeneration.

\subsection{Training Compute and Iteration Count}
\label{app:training-compute}
Optimizer updates measure learning progress but do not represent a fixed
computational budget. Each update follows image generation, visual critique, and acquisition filtering; with post-revision verification, accepted examples additionally require regeneration and an additional round of critique. When post-revision verification is required,
the measured checkpoint profiles in Table~\ref{tab:additional-compute} accept 20.20--34.48\% of acquisition attempts, corresponding to approximately three to five attempts per accepted experience. Rejected
candidates therefore contribute substantial overhead without increasing
the update count. Wall-clock cost depends on difficulty of training dataset, generation cost, and critic latency.

\begingroup
\subsubsection{Measured acquisition and training cost}
\label{app:additional-compute}
Table~\ref{tab:additional-compute} reports cumulative acquisition and
student-training costs at the measured checkpoints. Attempted counts
acquisition attempts; accepted counts experiences retained for OPSD after
the revised image passes post-revision assessment. GPU-hours are elapsed time multiplied
by the number of allocated GPUs. Acquisition costs include rejected
attempts, and GPU-hours per update divide the cumulative reported cost by
the number of optimizer updates.

\begin{table}[htbp]
\centering
\small
\setlength{\tabcolsep}{3.5pt}
\caption{Cumulative acquisition and student-training
costs. All costs are allocated GPU-hours. Component costs and totals are
reported separately; displayed totals differ from component sums by at
most 0.02 GPU-hours.}
\label{tab:additional-compute}
\begin{tabular}{rrrrrrrr}
\toprule
Update & Attempted & Accepted & Acceptance &
\shortstack{Acquisition\\GPU-h} & \shortstack{OPSD\\GPU-h} &
\shortstack{Total\\GPU-h} & \shortstack{GPU-h\\per update} \\
\midrule
\multicolumn{8}{c}{GenEval} \\
20 & 540 & 161 & 29.81\% & 21.15 & 1.11 & 22.26 & 1.11 \\
40 & 928 & 320 & 34.48\% & 37.04 & 2.22 & 39.26 & 0.98 \\
60 & 1,500 & 480 & 32.00\% & 58.20 & 3.34 & 61.54 & 1.03 \\
80 & 2,024 & 641 & 31.67\% & 76.86 & 4.45 & 81.32 & 1.02 \\
100 & 2,664 & 800 & 30.03\% & 99.80 & 5.55 & 105.36 & 1.05 \\
\midrule
\multicolumn{8}{c}{OCR} \\
20 & 792 & 160 & 20.20\% & 50.74 & 1.11 & 51.85 & 2.59 \\
40 & 1,480 & 320 & 21.62\% & 94.14 & 2.21 & 96.36 & 2.41 \\
60 & 2,208 & 480 & 21.74\% & 139.69 & 3.31 & 143.00 & 2.38 \\
80 & 2,968 & 640 & 21.56\% & 185.63 & 4.41 & 190.04 & 2.38 \\
100 & 3,748 & 801 & 21.37\% & 231.48 & 5.52 & 237.01 & 2.37 \\
\midrule
\multicolumn{8}{c}{GenEval2} \\
20 & 628 & 160 & 25.48\% & 39.52 & 1.12 & 40.65 & 2.03 \\
40 & 1,240 & 320 & 25.81\% & 79.62 & 2.23 & 81.85 & 2.05 \\
60 & 1,840 & 480 & 26.09\% & 118.00 & 3.33 & 121.35 & 2.02 \\
80 & 2,424 & 641 & 26.44\% & 156.61 & 4.44 & 161.06 & 2.01 \\
100 & 3,068 & 803 & 26.17\% & 203.80 & 5.64 & 209.46 & 2.09 \\
\bottomrule
\end{tabular}
\end{table}

The reported 150-update profile contains 4,188 acquisition attempts and
1,202 accepted experiences (28.7\%). Acquisition accounts for 154.29
GPU-hours and OPSD for 8.32 GPU-hours, totaling 162.61 GPU-hours or 1.08
GPU-hours per update. Experience acquisition therefore dominates the
reported adaptation cost.

\subsubsection{Inference latency}
\label{app:additional-latency}
Across the GenEval evaluation prompts, direct generation averages 4.08
seconds per prompt, compared with 50.55 seconds for the base model with
one reflection iteration, a latency ratio of approximately $12.4\times$.
The corresponding native GenEval scores are 0.818 for the adapted model
with direct generation and 0.826 for the base model with reflection.
Table~\ref{tab:additional-latency} separates the reflection components.
Merger and regeneration times are averaged over all evaluation prompts
when calculating the total; their conditional means apply only to prompts
for which those operations are invoked.

\begin{table}[htbp]
\centering
\small
\caption{Measured GenEval inference latency in seconds
per prompt. Conditional means cover only prompts on which the operation
is invoked; the all-prompt means sum to the reflection total.}
\label{tab:additional-latency}
\begin{tabular}{lrr}
\toprule
Phase & All prompts & When invoked \\
\midrule
Direct image generation & 4.08 & --- \\
\midrule
Initial generation in reflection run & 4.08 & --- \\
Critique & 39.70 & --- \\
Merger / prompt rewrite & 5.35 & 15.66 \\
Regeneration & 1.42 & 4.15 \\
\midrule
Total: direct inference & 4.08 & --- \\
Total: one reflection iteration & 50.55 & --- \\
\bottomrule
\end{tabular}
\end{table}
\endgroup

\subsection{Teacher Parameterization}
\label{app:teacher}
\begin{figure}[t]
\centering
\includegraphics[width=.68\linewidth]{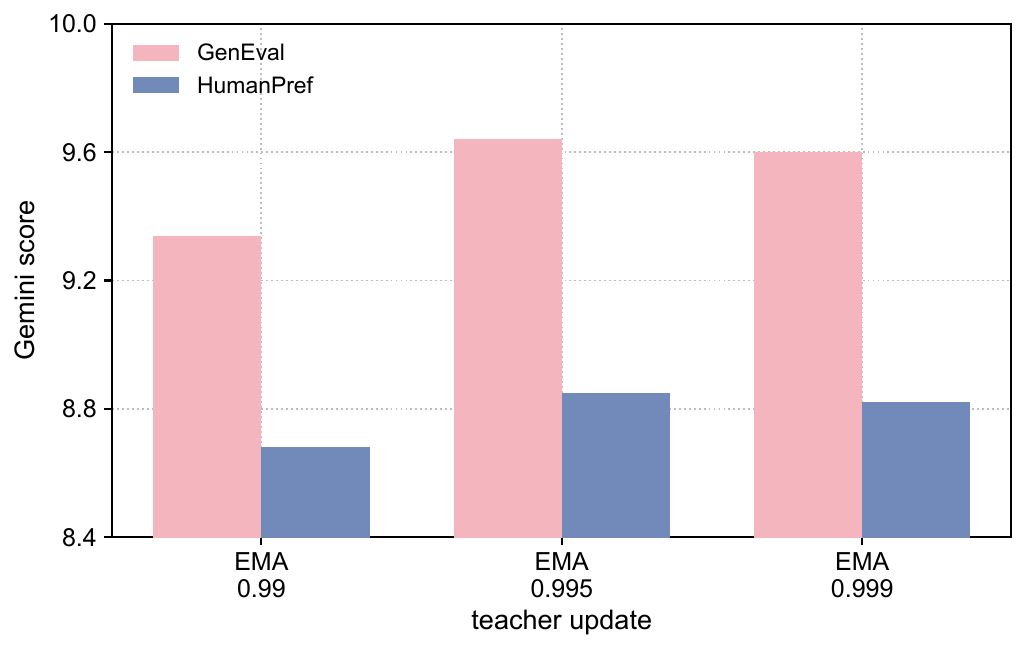}
\caption{\textbf{EMA teacher decay on GenEval.} Task and HumanPref scores across three EMA decays. Both metrics favor 0.995 and 0.999 over 0.99, with 0.995 achieving the highest scores in this sweep.}
\label{fig:teacher}
\end{figure}

The main runs use an EMA teacher with decay 0.999. Figure~\ref{fig:teacher} shows that slower teacher updates perform better within the tested range. Decreasing EMA further can slightly increase performance but risks becoming more unstable, as seen with 0.99. The benefit of other schedules, such as a time-varying teacher schedule, remains unexplored.

\section{Critique Interface}
\label{app:prompts}
The critic operates on the original prompt and the generated image. Its feedback is separated into semantic, text, and quality axes before synthesis. Semantic critique checks that requested objects exist with the correct number, attributes, and spatial relations. Text critique checks that specified words are rendered faithfully. Quality critique addresses visible defects in anatomy, geometry, layout, and rendering. A synthesis stage converts these observations into a prompt that can stand alone as a generation request.

The required behavior is to preserve the original intent, identify concrete failures, and express changes in terms that the generator can execute. A request without a visible problem can return \texttt{KEEP}. A revision must not simply refer to ``the previous image'' because the teacher is conditioned on text during generation. The following summary describes the interface; exact prompt templates are provided in the accompanying source file.

\begin{quote}\small
\textbf{Input:} the user's original generation prompt and the current draft image.\\
\textbf{Axis critique:} inspect semantic constraints, requested text, or visual quality; identify visible errors relevant to that axis.\\
\textbf{Synthesis:} combine actionable corrections into a self-contained generation prompt that preserves the user's request.\\
\textbf{No revision:} emit \texttt{KEEP} when the generation should remain unchanged.
\end{quote}

\paragraph{Why prompt synthesis matters.}
Raw criticism often contains negative descriptions of the current output. A generation model instead needs a coherent specification of the desired output. The synthesis stage resolves this mismatch: it combines corrections and restates what should be present in the regenerated scene. Teacher conditioning consequently carries both the original task and the information gained by inspecting the failure. The student does not receive this additional context, making its successful imitation an actual transfer of the correction into the original-prompt policy.

\paragraph{Why match on student states?}
Teacher-generated images may occupy trajectories the student rarely visits. Matching both models at the student's noisy states supplies local guidance where the student currently needs it. It also separates two questions: which correction the critic proposes, and how that correction changes the generator's next transition. Equation~\ref{eq:loss} uses the latter as a dense supervision signal. For deterministic flow sampling, this is an $L_2$ transition objective; no stochastic-policy KL interpretation is required for the implementation used here.

\section{Datasets}
\label{app:datasets}

We study compositional image generation with GenEval and GenEval2, and
visual text rendering with an OCR prompt collection. These tasks provide
textual specifications for image generation. Training uses generated
drafts and corrective feedback, without paired target images. Below, we
describe the benchmark contents, their use in our experiments, and the
HumanPref measure used to assess visual quality across tasks.
\appendixname~\ref{app:eval} specifies the shared evaluation protocol.

\paragraph{GenEval.} GenEval~\citep{geneval} evaluates whether a text-to-image model can translate explicit object-level requirements into an image. Its standard
evaluation set contains 553 prompts across six categories: single object
(80), two objects (99), counting (80), colors (94), position (100), and
color attribute binding (100). The prompts use controlled combinations of
object names, quantities, colors, and spatial relations, enabling errors
to be attributed to specific compositional skills. Attribute binding,
for example, requires each requested color to be assigned to the correct
object, rather than merely appearing somewhere in the image. Training prompts are drawn from GenEval-format training metadata. 

\paragraph{GenEval2.} GenEval2~\citep{geneval2} provides a harder compositional evaluation,
introduced to address the saturation and evaluator drift observed on
GenEval. We additionally use its newer prompt suite to reduce reliance on
the extensively reused GenEval benchmark and potential effects of
benchmark-specific optimization or data contamination. Its prompts combine
more simultaneous requirements, making successful generation more
demanding. The benchmark contains 800 prompts, with
100 prompts at each atomicity level from 3 to 10. An \emph{atom} is an
individual semantic requirement, such as the presence of an object, an
attribute assigned to it, a count, or a relation between objects;
\emph{atomicity} measures how many such requirements are combined in a
prompt. Its vocabulary covers 40 objects, 18 attributes, and nine
relations, including spatial and action relations. The released annotations
support analysis by compositional complexity and by the skills needed to
satisfy a prompt. This makes GenEval2 useful for testing whether feedback
helps the generator satisfy several interacting constraints. Our training stream uses a separate compositional training collection,
while the main holistic evaluation uses all 800 benchmark prompts with fixed
prompt--seed assignments.

\paragraph{OCR: Visual Text Rendering.} The OCR task evaluates the ability to render specified text inside a generated image, following the visual text-rendering setting used by
Flow-GRPO~\citep{flowgrpo}. Each prompt describes a scene or surface and
includes a quoted target string that should appear in the image.
Successful generation requires both readable lettering and faithful
reproduction of the requested text within the described scene. The source
collection comprises approximately 20,000
training prompts and 1,018 test prompts. Its scene-conditioned requests
cover contexts such as signs, posters, and product labels.

\begingroup
\paragraph{Training--evaluation separation.}
For GenEval2, we construct 50,000 unique adaptation prompts and a separate 800-prompt development set, reserving the official 800-prompt benchmark for evaluation. Construction excludes candidates matching an official or previously accepted example by normalized prompt text or canonical annotated-scene identity, including objects, counts, attributes, and relations. Scene comparison canonicalizes entity order, symmetric relations, and inverse-equivalent spatial relations. For GenEval, the training collection has zero exact or normalized prompt overlap with the 553 evaluation prompts. For OCR, we use the released Flow-GRPO splits of 19,653 training rows and 1,018 evaluation prompts.
\endgroup

\section{Evaluation Definitions and Additional Results}
\label{app:eval}
We distinguish three scoring protocols: Gemini judgments of individual
semantic checks, holistic Gemini ratings, and native benchmark metrics.
Using an official prompt set does not determine which evaluator produced
the score. All scores described here are evaluation measures. UniEvo-VL
uses corrective feedback and distillation without training rewards;
evaluation scores do not enter its training objective.

\paragraph{Evaluation inputs and pairing.}
Each generated image is assessed against the original benchmark prompt.
For inference with corrective feedback, the final image is still judged
against that original request; the revised generation prompt does not
replace the evaluation target. Direct generation and generation with
feedback are reported separately. Corresponding prompts and generation
seeds are paired within each experiment across checkpoints and inference conditions, but not across the earlier verified and unverified Qwen-Image-2512 configurations. The matched Qwen-Image-2.1 controls are described in Appendix~\ref{app:qwen21-controls}.

The Qwen-Image-2512 full-cohort evaluations use $1024\times1024$ images, 20 denoising
steps, and a true classifier-free guidance scale of 4.0. The verified recipe uses generation
seed $42 + 20{,}000{,}000 + i$, whereas the unverified full-set runs use $20{,}000{,}000+i$, where $i$ is the zero-based metadata-row
index. Repeated GenEval prompts occupy distinct rows and receive distinct
seeds. Table~\ref{tab:full-evaluation-protocols} specifies the full cohorts.
The main holistic learning curves use 553 GenEval, 800 GenEval2, and 1,018 OCR
prompts, with one image per prompt. The generation--reflection/model-gap study
uses 553 unique GenEval prompts. Those analyses retain their own sampling
settings and are not pooled with the full-cohort evaluations.

\begin{table}[htbp]
  \centering
  \small
  \caption{Full-cohort Gemini evaluation for the verified recipe. Image counts are per checkpoint
  and inference condition. HumanPref is evaluated separately on each
  corresponding image pool.}
  \label{tab:full-evaluation-protocols}
  \begin{tabular}{@{}lrrl@{}}
    \toprule
    Task & Unique prompts & Images & Task-scoring protocol \\
    \midrule
    GenEval & 553 & 2,212 & 8,392 binary checks \\
    GenEval2 & 800 & 800 & 6,012 binary checks \\
    OCR & 1,018 & 1,018 & Edit distance \\
    \bottomrule
  \end{tabular}
\end{table}

\subsection{Gemini Evaluation}
\label{app:gemini-evaluation}

The judge settings, request rubrics, and failure policy below specify the
full-cohort scoring implementation. The full-set holistic learning curves and historical subset
analysis retain their experiment-specific evaluation settings.

\paragraph{Shared judge configuration.}
We use \texttt{gemini-2.5-flash}, with temperature 0 and judge seed 42.
Each request contains the image followed by text specifying the benchmark,
original prompt, and evaluation instructions. Responses are constrained
to a JSON schema. The judge seed is separate from the generation seed;
these settings reduce sampling variability without guaranteeing identical
responses across repeated calls to the hosted model.

\paragraph{Atom-based semantic evaluation.}
We construct the checks from benchmark annotations before judging the
image. For GenEval, the metadata determine object-presence and exact-count
checks, with color and spatial-relation checks where specified. For
GenEval2, we use every released VQA question, expected answer, and skill
label. The skills cover objects, attributes, counts, spatial relations,
and actions. Auxiliary questions, including count questions whose answer
is ``one,'' remain part of scoring even when they do not contribute to
the benchmark's annotated atomicity.

One semantic request contains all checks for an image. Each check includes
an identifier, skill, question, and expected answer. Gemini is instructed
to judge each check independently from visible evidence, to treat expected
answers as targets rather than evidence, and to return \texttt{uncertain}
when the image is ambiguous. Its response must contain exactly one
\texttt{atom\_id}, \texttt{verdict}, and nonempty \texttt{evidence} entry
for each requested check. Verdicts are \texttt{pass}, \texttt{fail}, or
\texttt{uncertain}; duplicate, missing, or extra identifiers invalidate
the response.

Let $N$ denote the number of evaluated image instances, $K_i$ the number
of checks for image $i$, and $z_{ij}=1$ only when check $j$ receives
\texttt{pass}. We report
\begin{align}
S_{\mathrm{check}} = \frac{\sum_{i=1}^{N}\sum_{j=1}^{K_i} z_{ij}}{\sum_{i=1}^{N}K_i}, \quad S_{\mathrm{strict}} = \frac{1}{N}\sum_{i=1}^{N}\mathbf{1}\!\left[\sum_{j=1}^{K_i}z_{ij}=K_i\right]
    \label{eq:gemini-strict-accuracy}
\end{align}
Check accuracy weights every check equally. Strict success requires all
checks for an image to pass. For full GenEval2 evaluation, their
denominators are 6,012 checks and 800 images, respectively. For verified GenEval,
strict success is averaged over 2,212 images, rather than counting a
prompt as successful if any of its four images passes. Unverified GenEval uses 553 images and 2,098 checks. We also retain
per-skill check accuracy and strict success grouped by GenEval2 atomicity.

\paragraph{Holistic task ratings and HumanPref.}
Holistic evaluation uses one task-specific request per image. The
composition rubric is ``Score compositional fidelity to the prompt on an
integer 0--10 scale.'' The OCR rubric substitutes ``rendered text
fidelity'' for ``compositional fidelity.'' Each response supplies an
integer \texttt{score} in $[0,10]$ and a short \texttt{rationale}.
These ratings assess the image as a whole and are not calculated by
rescaling atom accuracy. The OCR rating is a model judgment of text
rendering, rather than an OCR edit-distance metric.

In both semantic modes, a separate request scores HumanPref using the
rubric ``Score overall image quality and human preference on an integer
0--10 scale.'' We report the arithmetic mean of each scalar measure
separately over its valid returned ratings. HumanPref is an absolute
automated proxy; it is neither a human annotation study nor a pairwise
win rate, and it is not combined with semantic correctness.

\paragraph{Uncertainty and failed requests.}
Retryable service failures receive up to three retries after the initial
request. In atom mode, uncertain checks earn no credit. A failed semantic
request also earns no credit, while all its scheduled checks and its image
remain in the full-evaluation denominators. Such failures are recorded
separately from valid negative judgments. A failed HumanPref request in
atom mode yields a missing scalar rating and is excluded from the
HumanPref mean. Full holistic scoring instead stops on an unrecoverable
request; a partially scored cohort is not silently reported as complete.
We preserve individual verdicts, rationales, and error records alongside
the aggregate scores. Particular requests are also rejected without specific safety reasons provided by the API.

\subsection{Native Benchmark Metrics}
\label{app:native-evaluation}

\paragraph{Official GenEval.} We use the official scorer for all native GenEval (GenEval1) results.
The official evaluator~\citep{geneval} uses Mask2Former detections,
CLIP-based color classification, and geometric rules for spatial
relations. Each image receives a binary correctness label according to
the benchmark metadata. The overall score is the unweighted mean of the
six category accuracies:
\begin{equation}
  S_{\mathrm{GenEval}} = \frac{1}{6}\sum_{c=1}^{6}
  \frac{1}{N_c}\sum_{i\in c} b_i,
  \label{eq:official-geneval}
\end{equation}
where $b_i$ is the official binary label and $N_c$ is the number of images
in category $c$. This category-balanced metric differs from both Gemini
check accuracy and Gemini strict success. Results labeled official
GenEval refer to this detector-based protocol.

\paragraph{Native GenEval2 reference.}
GenEval2's released Soft-TIFA evaluator~\citep{geneval2} uses
Qwen3-VL-8B-Instruct to obtain soft answer probabilities for the VQA
checks. Its prompt-level score is the geometric mean of those
probabilities, averaged across prompts. For question probabilities
$p_{ij}$, the geometric-mean (GM) and arithmetic-mean (AM) summaries are
\begin{equation}
  S_{\mathrm{GM}} = \frac{1}{N}\sum_i
       \left(\prod_{j=1}^{K_i}p_{ij}\right)^{1/K_i},
  \qquad
  S_{\mathrm{AM}} = \frac{1}{N}\sum_i
       \frac{1}{K_i}\sum_{j=1}^{K_i}p_{ij}.
  \label{eq:soft-tifa-definitions}
\end{equation}
The released evaluator reports both measures multiplied by 100. AM gives each prompt equal weight, unlike the pooled Gemini check
accuracy. Our main GenEval2 table reports atomic Gemini ratings and native Soft-TIFA over
all 800 official prompts; Gemini binary atom judgments are a separate supplementary protocol.

\paragraph{Flow-GRPO OCR reference.}
The Flow-GRPO reference evaluator~\citep{flowgrpo} uses English PaddleOCR
to read the image and compares the recognized text with the first
double-quoted target string in the prompt. It concatenates recognized
text regions with positive recognition confidence, removes ASCII spaces,
and lowercases both strings. A
substring match receives 1; otherwise its score is
$1-\min(d_{\mathrm{Lev}}(\hat y,y),|y|)/|y|$, where $y$ is the normalized
target, $\hat y$ is the recognized text, and $d_{\mathrm{Lev}}$ is
Levenshtein edit distance. Recognition exceptions receive zero, and scores
are averaged over all images. This
recognition-based reference is distinct from the Gemini OCR ratings
reported in our task evaluations.

\paragraph{Interpreting the scores.}
Detector errors can cause official GenEval to penalize images that
satisfy the prompt, consistent with the evaluator drift documented by
\citet{geneval2}. Because our reward-free training does not optimize these
scores directly, improved prompt fulfillment need not produce a
near-perfect detector score. Gemini judgments also remain automated
assessments. We interpret each metric according to its own evaluator,
aggregation rule, and image population, alongside qualitative evidence.

\subsection{Comparison with other methods}

\begin{table}[H]
\caption{\textbf{GenEval performance on 553 prompts.}
Scores use the official evaluator (0--1; higher is better).
Our update-200 rows use training without post-revision verification;
update-140 rows use training with post-revision verification.
Each checkpoint is evaluated with direct generation and with one
critique iteration.
The Qwen-Image-2512 and update-200 rows use one evaluated image per
prompt; update-140 rows use four images per prompt (2,212 images).
Bold and underlined values indicate the highest and second-highest
scores in each column, with ties at the displayed precision.}
\label{tab:geneval}
\centering\small
\setlength{\tabcolsep}{3.8pt}
\begin{tabular*}{\linewidth}{@{\extracolsep{\fill}}lrrrrrrr@{}}
\toprule
Method & Single & Two & Count & Color & Position & Attr. & Overall\\
\midrule
SD3.5-L
& 0.98 & 0.89 & 0.73 & 0.83 & 0.34 & 0.47 & 0.71\\
HiDream-I1
& \textbf{1.00} & \textbf{0.98} & 0.79 & \underline{0.91}
& 0.60 & 0.72 & 0.83\\
Z-Image
& \textbf{1.00} & 0.94 & 0.78 & \textbf{0.93}
& 0.62 & \textbf{0.77} & 0.84\\
SANA-1.5
& \underline{0.99} & 0.93 & \textbf{0.86} & 0.84
& 0.59 & 0.65 & 0.81\\
LongCat
& \underline{0.99} & \textbf{0.98} & \textbf{0.86} & 0.86
& 0.75 & \underline{0.73} & \textbf{0.87}\\
BAGEL
& \underline{0.99} & 0.94 & 0.81 & 0.88
& 0.64 & 0.63 & 0.82\\
\midrule
Qwen-Image-2512
& \textbf{1.00} & 0.93 & 0.61 & 0.89
& 0.46 & 0.58 & 0.75\\
\ours\ (200, direct)
& \underline{0.99} & \underline{0.97} & 0.70 & 0.86
& 0.68 & 0.65 & 0.81\\
\quad +1 critique (200)
& \underline{0.99} & \textbf{0.98} & 0.75 & 0.89
& \underline{0.78} & 0.69 & \underline{0.85}\\
\ours\ + verification (140, direct)
& \underline{0.99} & 0.96 & 0.74 & 0.85
& 0.77 & 0.60 & 0.82\\
\quad +1 critique (140)
& \underline{0.99} & 0.96 & \underline{0.82} & 0.89
& \textbf{0.81} & 0.62 & \underline{0.85}\\
\bottomrule
\end{tabular*}
\end{table}

Table~\ref{tab:geneval} shows that \ours\ raises Qwen-Image-2512 to competitive GenEval
performance without using external evaluator signal during training.
Direct generation reaches 0.81 without post-revision verification
and 0.82 with verification, compared with the reported Qwen-Image-2512
baseline of 0.75. One critique iteration raises both configurations
to 0.85, the second-highest overall score among the listed methods.
The verified configuration matches BAGEL's reported overall score
in direct generation and reaches the highest spatial-position score
(0.81) with one critique iteration. These results should be interpreted
in light of the training signal available to our method. Approaches
that optimize native evaluator-derived rewards receive feedback
directly aligned with the benchmark's scoring criteria. Our adaptation
forgoes this advantage: it relies on semantic diagnoses and corrections
from a critic, which must transfer to the native evaluator's object,
count, attribute, and spatial checks. The resulting native-score
improvements therefore provide evidence that critic-derived semantic
supervision can improve compositional generation under a separate
evaluation mechanism, without directly optimizing its reward.

\subsection{Full judge-less learning curve}
\label{app:no-judge-learning-curve}

\begin{table}[t]
\caption{Full-set learning curves without post-revision verification, rounded to three decimals. Each checkpoint uses 553 GenEval, 800 GenEval2, and 1,018 OCR prompts; task and HumanPref scores retain their 0--10 scale.}
\label{tab:judgeless-learning-curve}
\centering\small
\begin{tabular}{rrrrrrr}
\toprule
 & \multicolumn{2}{c}{GenEval} & \multicolumn{2}{c}{GenEval2} & \multicolumn{2}{c}{OCR}\\
Step & Task & HumanPref & Task & HumanPref & Task & HumanPref\\
\midrule
0 & 8.958 & 8.503 & 6.228 & 6.716 & \textbf{9.402} & \textbf{9.074}\\
40 & 9.128 & 8.526 & 6.277 & 6.726 & 9.307 & 9.025\\
80 & 9.134 & 8.544 & 6.321 & 6.737 & 9.364 & 9.054\\
120 & 9.316 & 8.671 & 6.356 & 6.824 & 9.301 & 9.042\\
160 & 9.421 & 8.759 & 6.320 & 6.732 & 9.377 & 9.064\\
200 & \textbf{9.573} & \textbf{8.787} & \textbf{6.451} & \textbf{6.856} & 9.372 & 9.033\\
\bottomrule
\end{tabular}
\end{table}

Table \ref{tab:judgeless-learning-curve} motivate post-revision verification as a safeguard
for training-signal reliability. Although GenEval improves steadily,
GenEval2 exhibits an intermediate decline, and OCR fluctuates. These trajectories suggest that
critique-guided prompt revision without verification may not consistently provide
beneficial supervision on relatively more difficult-to-learn tasks. A primary cause of this is that producing a revised prompt does not guarantee
that the regenerated image resolves the original error. Post-revision
verification checks this outcome against the original request before
accepting the revision for training. By admitting confirmed corrections
and filtering unsuccessful revisions, this step is designed to provide
a more reliable training signal and support more stable learning.

\subsection{Historical Difficulty Analysis on 500-Prompt Subsets}
\label{app:historical-difficulty}
These historical analyses use 500 prompts per task and are not full-set results. Easy contains baseline scores 5--10 and Hard contains 0--4; membership is fixed before training. GenEval uses update 180 and the other tasks use update 200. The Hard subsets contain 51, 144, and 24 prompts, respectively.

\paragraph{Normalized difficulty gains.}
For a fixed subset $B$ and a task score $s\in[0,10]$, we report
\begin{equation}
\Delta_B^{(\%)}=100\left(\frac{\E[s_{\rm evolved}\mid B]-\E[s_{\rm base}\mid B]}{10}\right).
\end{equation}
These are percentage-point changes on a normalized 0--100 score. The denominator is the full score range, not the baseline mean. Subsets are defined once from the baseline, with no regrouping after training. Values in Figure~\ref{fig:historical-difficulty} are rounded to the nearest integer. The small size of OCR's Hard subset should be considered when interpreting its large normalized gain.

\subsection{Native GenEval2 Soft-TIFA Across Updates}
\label{app:soft-tifa}
All points use 800 prompts and the released Qwen3-VL-8B-Instruct VQA evaluator. GM is the primary native metric; AM is a supplementary arithmetic-mean summary. Both retain the evaluator's 0--100 scale. The main result uses update 200, without selecting the later peak at update 220.
\begin{figure}[t]
\centering
\includegraphics[width=\linewidth]{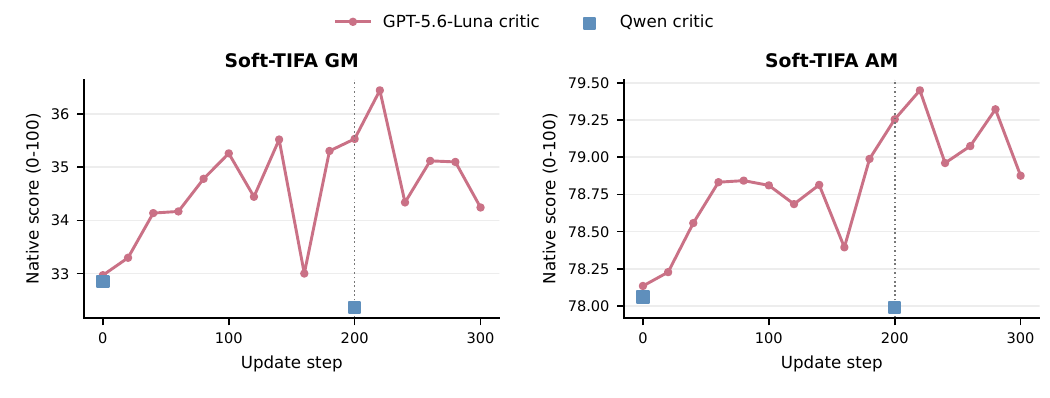}
\caption{Complete GenEval2 Soft-TIFA trajectory for GPT-5.6-Luna critique, updates 0--300. The Qwen-critic baseline and update-200 endpoints are shown as separate markers. GM and AM use each run's own baseline.}
\label{fig:soft-tifa}
\end{figure}
\begin{table}[t]
\centering\small
\caption{Paired native GenEval2 results at update 200 (800 prompts). Deltas use unrounded scores.}
\begin{tabular}{lrrrrrr}
\toprule
 & \multicolumn{3}{c}{Soft-TIFA GM} & \multicolumn{3}{c}{Soft-TIFA AM}\\
Critic & Base & Evolved & $\Delta$ & Base & Evolved & $\Delta$\\
\midrule
Qwen-VL & 32.86 & 32.37 & $-$0.49 & 78.06 & 77.99 & $-$0.07\\
GPT-5.6-Luna & \textbf{32.97} & \textbf{35.53} & \textbf{+2.56} & \textbf{78.13} & \textbf{79.25} & \textbf{+1.12}\\
\bottomrule
\end{tabular}
\end{table}

\subsection{GenEval Evaluator Limitations}
\label{app:geneval-drift}

GenEval relies on object detection, color classification, and geometric
rules, whose errors can penalize images that satisfy the prompt.
Consistent with this limitation,
\citet{geneval2} report 96.7\% human-assessed correctness for
Gemini~2.5 Flash Image with prompt rewriting, compared with an automatic
GenEval score of 82.1\%. UniEvo-VL uses corrective feedback and
distillation without training rewards; GenEval scores are used for
evaluation and do not enter the training objective. Consequently,
improved prompt fulfillment need not translate into near-perfect
GenEval scores, and a score below 0.95 should not automatically be
interpreted as a corresponding deficit in semantic correctness.
We therefore assess progress jointly through GenEval, the harder
GenEval2 benchmark, and qualitative inspection.

\subsection{Unreturned Judge Responses}
\label{app:unreturned-responses}

Few generated-image instances lack their Gemini task judgment and
their visual-preference rating. A
diagnostic request for each of these ten judgments (over all experiments) returned no candidate
and no parsable answer; the API reported
\texttt{prompt\_feedback.block\_reason=OTHER} with no safety-category
ratings. This response does not establish that the generated image
violated the prompt or identify a specific reason for the block. We
therefore distinguish an unreturned judgment from a valid negative
judgment. 

\subsection{Image Comparison}
\begin{figure}[t]
    \centering
    \includegraphics[width=\linewidth]{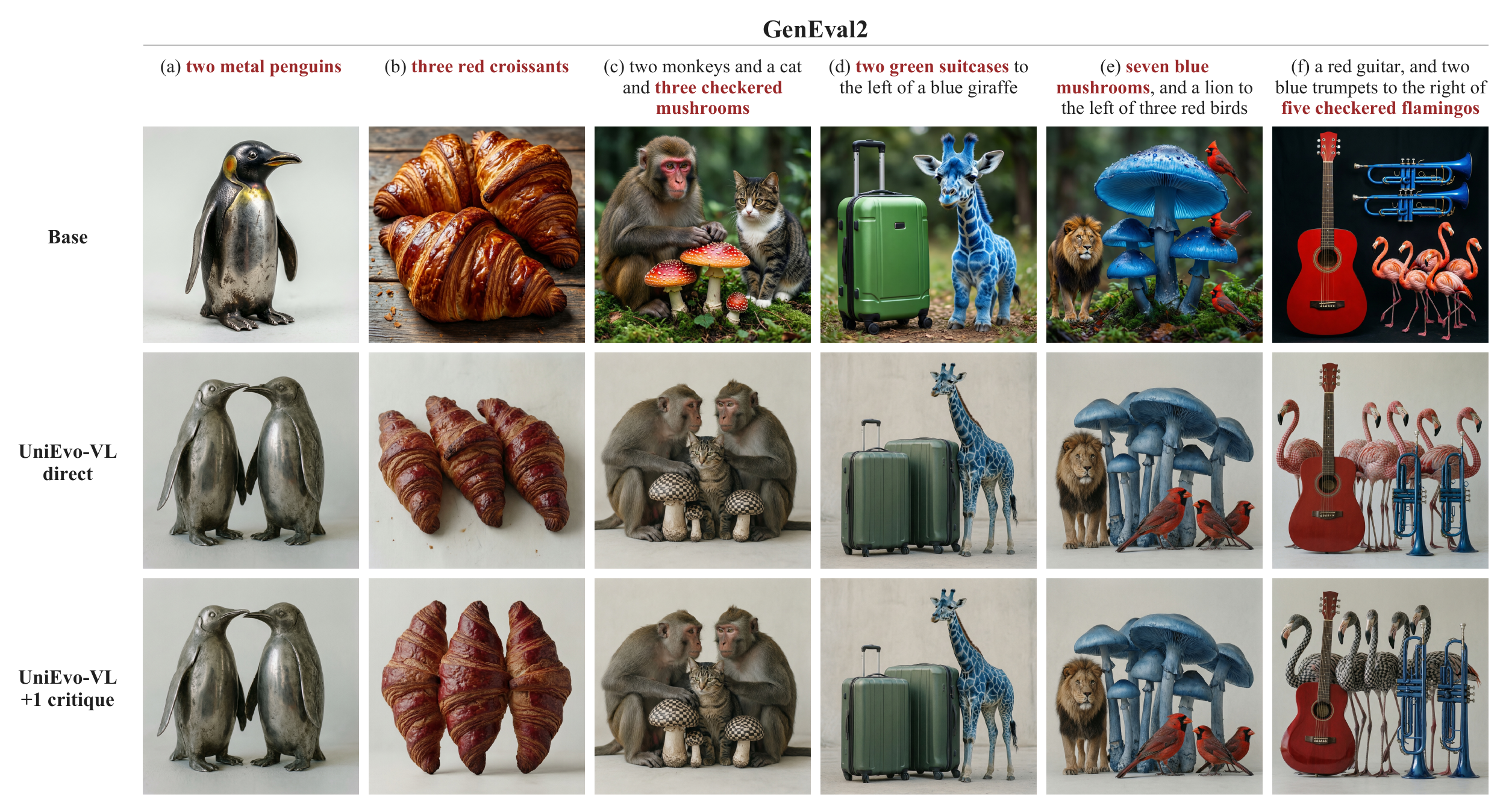}
    \caption{\textbf{Qualitative examples of UniEvo-VL on GenEval2.}
    Examples cover object counts, colors, materials, and patterns.
    Rows show the base model, UniEvo-VL direct generation, and
    UniEvo-VL with up to one critique.
    Each column shares the original prompt and seed; both UniEvo-VL
    rows use checkpoint 160 of the Qwen critic-confirmed variant.
    A \texttt{KEEP} decision retains the direct image.}
    \label{fig:qualitative-geneval2}
\end{figure}

\begin{figure}[t]
    \centering
    \includegraphics[width=\linewidth]{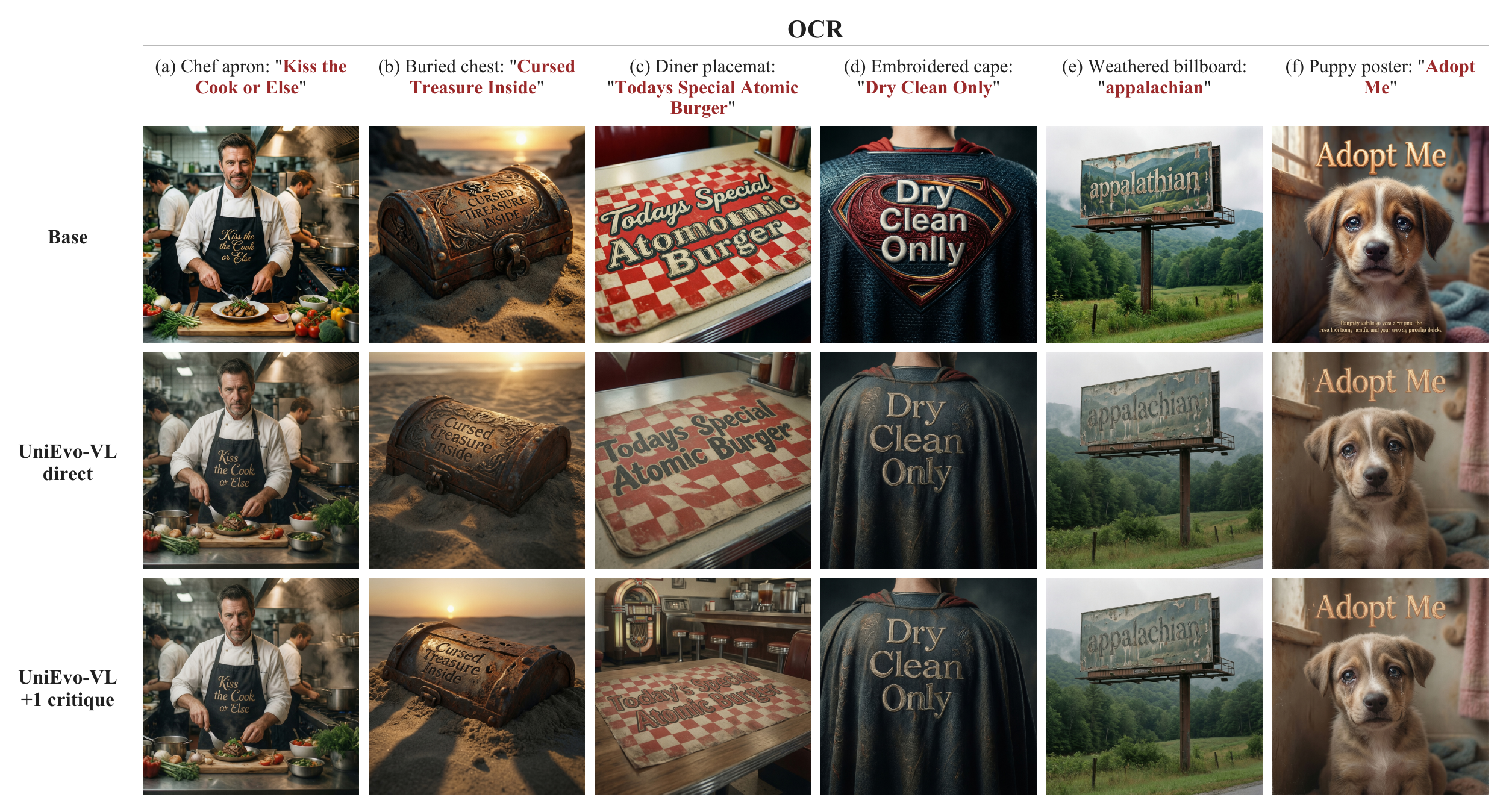}
    \caption{\textbf{Qualitative examples of UniEvo-VL on OCR.}
    Examples illustrate text rendering across varied scenes;
    headings summarize the scene and quote the requested text.
    Rows show the base model, UniEvo-VL direct generation, and
    UniEvo-VL with up to one critique.
    Each column shares the original prompt and seed; both UniEvo-VL
    rows use checkpoint 130 of the Qwen critic-confirmed variant.
    A \texttt{KEEP} decision retains the direct image.}
    \label{fig:qualitative-ocr}
\end{figure}

\paragraph{Influence of feedback-prompt design.}
Some outputs on GenEval and GenEval2 exhibit muted colors and
frequent plain, light-colored backgrounds. This pattern is consistent
with a stylistic bias introduced by our feedback-prompt design for realism. Although this aesthetic differs from more common high-saturation AI-generated photos, it scores higher on our selected human preference benchmarks. This motivates greater caution when designing aesthetic prompts during feedback for self-evolving models as small biases could accumulate.

\begingroup
\section{Additional GenEval2 Controls}
\label{app:qwen21-controls}

\begingroup
\paragraph{Results with Qwen-Image-2.1.}
Corrective adaptation also improves the newer Qwen-Image-2.1 backbone on GenEval2 (Table~\ref{tab:qwen21-full}). Across seeds 42, 43, and 44, direct-generation Soft-TIFA GM rises from $13.30\pm1.04$ to $40.37\pm0.89$ at update 150; the gain is positive in every seed. The mean within-seed gain is $27.07\pm1.49$ GM points and $20.86\pm1.07$ AM points. One additional reflection raises the final scores to $50.27\pm1.43$ GM and $83.06\pm0.83$ AM.

\begin{table}[htbp]
\centering\small
\caption{Qwen-Image-2.1 on official GenEval2: all evaluated checkpoints. Soft-TIFA GM and AM are multiplied by 100; entries give mean $\pm$ sample SD across seeds 42, 43, and 44. Each endpoint uses 800 images and 6,012 checks. Direct generation uses the original prompt; +1 refl. adds one correction pass. Bold marks the final checkpoint.}
\label{tab:qwen21-full}
\begin{tabular*}{\linewidth}{@{\extracolsep{\fill}}rrrrr@{}}
\toprule
Update & Direct GM & Direct AM & +1 refl. GM & +1 refl. AM\\
\midrule
0 & $13.30\pm1.04$ & $58.61\pm0.60$ & $36.89\pm1.43$ & $74.82\pm0.61$\\
20 & $23.33\pm2.00$ & $69.65\pm0.85$ & $38.98\pm2.70$ & $77.56\pm1.21$\\
40 & $29.07\pm0.81$ & $73.44\pm0.75$ & $44.23\pm0.28$ & $80.13\pm0.24$\\
60 & $31.85\pm0.92$ & $75.14\pm0.60$ & $45.15\pm1.92$ & $80.66\pm0.47$\\
80 & $35.25\pm3.25$ & $76.82\pm2.02$ & $47.41\pm4.07$ & $81.79\pm1.73$\\
100 & $35.71\pm1.08$ & $77.16\pm0.08$ & $47.56\pm1.17$ & $81.95\pm0.26$\\
120 & $38.74\pm1.61$ & $78.85\pm0.83$ & $49.60\pm1.06$ & $82.95\pm0.37$\\
140 & $37.92\pm1.37$ & $78.52\pm0.68$ & $48.10\pm1.66$ & $82.18\pm0.16$\\
150 & $\mathbf{40.37\pm0.89}$ & $\mathbf{79.46\pm0.52}$ & $\mathbf{50.27\pm1.43}$ & $\mathbf{83.06\pm0.83}$\\
\bottomrule
\end{tabular*}
\end{table}
\endgroup

\paragraph{Qwen-Image-2.1 evaluation protocol.}
Table~\ref{tab:qwen21-full} reports nine checkpoints for each of three runs, with direct and one-reflection inference. Seed 42 includes its earlier training continuation; seeds 43 and 44 start fresh. The sample standard deviation summarizes these three training-and-generation trajectories. Each of the 54 endpoints uses the official 800-prompt cohort and 6,012 semantic checks, scored by the native GenEval2 Soft-TIFA evaluator. Update 150 is the configured training endpoint. Direct generation evaluates saved students on original prompts without inference-time reflection.

\subsection{Prompt-only rewriting versus corrective feedback}
\label{app:rewrite-full}
We distill conditioning from the official Qwen image prompt enhancer to test how much improvement ordinary prompt rewriting provides. The enhancer sees the original request without a generated draft or image critique; its rewritten text conditions the teacher while the student receives the original request. Both students are evaluated on the same official prompt cohort and generation seeds. Table~\ref{tab:rewrite-full} reports the complete seed-43 trajectory.

Feedback outperforms prompt-only rewriting at every evaluated post-training checkpoint. At update 150, the gap is $6.98$ GM and $6.33$ AM points. Step 0 evaluates the unadapted students using original prompts, before distillation. 

\begin{table}[htbp]
\centering\small
\caption{Prompt-only rewriting and corrective feedback, seed 43. Official GenEval2 Soft-TIFA $\times100$; 800 images and 6,012 checks per checkpoint. Both students generate directly from original prompts. Bold marks feedback's higher score in each post-training comparison.}
\label{tab:rewrite-full}
\begin{tabular*}{\linewidth}{@{\extracolsep{\fill}}rrrrr@{}}
\toprule
Update & Rewrite GM & Feedback GM & Rewrite AM & Feedback AM\\
\midrule
0 & 13.45 & 13.27 & 59.30 & 59.14\\
20 & 12.31 & \textbf{25.62} & 58.35 & \textbf{70.59}\\
40 & 17.91 & \textbf{29.94} & 62.39 & \textbf{74.04}\\
60 & 18.53 & \textbf{32.14} & 63.70 & \textbf{75.61}\\
80 & 21.76 & \textbf{36.51} & 67.02 & \textbf{77.89}\\
100 & 27.26 & \textbf{35.31} & 69.91 & \textbf{77.26}\\
120 & 29.23 & \textbf{36.91} & 70.61 & \textbf{78.02}\\
140 & 31.82 & \textbf{36.86} & 73.29 & \textbf{78.17}\\
150 & 32.39 & \textbf{39.37} & 72.53 & \textbf{78.86}\\
\bottomrule
\end{tabular*}
\end{table}

\subsection{Matched verification ablation}
\label{app:verification-control}
To separate post-revision verification from the optimization changes in the earlier Qwen-Image-2512 experiments, we repeat seed-43 Qwen-Image-2.1 training with the same optimizer settings, acquisition seeds per prompt, and evaluation protocol. The unverified run accepts well-formed corrective prompts directly, removing corrected-image regeneration and its acceptance test during acquisition. Initial \texttt{KEEP} responses and invalid corrections remain excluded, and teacher-conditioned distillation is unchanged. The resulting acquisition costs and accepted examples differ by design.

Table~\ref{tab:verification-selected} shows that without verification, direct-generation GM improves from 13.45 to 37.94, a gain of 24.49 points. Verification reaches 39.37 GM and 78.86 AM at update 150, versus 37.94 GM and 78.52 AM without verification. Both variants improve substantially in this matched seed, with a final advantage of 1.43 GM and 0.34 AM points for verification. The advantage is not uniform across training: without verification, GM is higher at updates 100 (37.27 versus 35.31) and 140 (39.43 versus 36.86). The value of verification may depend on the model and task; this single-seed comparison establishes its effect for the tested Qwen-Image-2.1 configuration.

\begin{table}[htbp]
\centering\small
\caption{Verification ablation on Qwen-Image-2.1, seed 43,
at selected checkpoints. Official GenEval2 Soft-TIFA
$\times100$, with 800 images and 6,012 checks per checkpoint.
Both variants use direct generation. Bold marks the higher
post-training score within each metric and row.}
\label{tab:verification-selected}
\begin{tabular*}{\linewidth}{@{\extracolsep{\fill}}rrrrr@{}}
\toprule
& \multicolumn{2}{c}{GM} & \multicolumn{2}{c}{AM}\\
Update & Unverified & Verified & Unverified & Verified\\
\midrule
0   & 13.45 & 13.27          & 59.30 & 59.14\\
40  & 27.69 & \textbf{29.94} & 71.72 & \textbf{74.04}\\
80  & 33.21 & \textbf{36.51} & 75.74 & \textbf{77.89}\\
120 & 36.36 & \textbf{36.91} & 77.21 & \textbf{78.02}\\
150 & 37.94 & \textbf{39.37} & 78.52 & \textbf{78.86}\\
\bottomrule
\end{tabular*}
\end{table}

\subsection{Configuration across backbones}
\label{app:qwen21-config}
The Qwen-Image-2.1 runs retain the optimization recipe used for verified Qwen-Image-2512 (Table~\ref{tab:backbone-config}). The sampling guidance changes from CFG 4 to CFG 1 because the official Qwen-Image-2.1 pipeline is intended for guidance-free sampling.\footnote{Official Diffusers Qwen-Image-2.1 implementation: \url{https://github.com/huggingface/diffusers/blob/main/src/diffusers/pipelines/qwenimage21/pipeline_qwenimage21.py}.} CFG 1 is used consistently for the Qwen-Image-2.1 baseline, feedback runs, and prompt-only control. The earlier unverified Qwen-Image-2512 GenEval2 configuration instead used learning rate $10^{-4}$ and eight acquisition seeds per prompt; its comparison with the verified recipe changes these settings together with verification. The within-Qwen-Image-2.1 ablation above holds them fixed.

\begin{table}[htbp]
\centering\small
\caption{Optimization and sampling settings for verified Qwen-Image-2512 and the Qwen-Image-2.1 feedback runs on GenEval2. The seed-43 verification ablation holds the Qwen-Image-2.1 column fixed while removing post-revision verification and selection.}
\label{tab:backbone-config}
\setlength{\tabcolsep}{4pt}
\begin{tabular*}{\linewidth}{@{\extracolsep{\fill}}lcc@{}}
\toprule
Setting & Qwen-Image-2512 & Qwen-Image-2.1\\
\midrule
Optimizer & AdamW & AdamW\\
Learning rate & $3\times10^{-4}$ & $3\times10^{-4}$\\
Resolution & $1024\times1024$ & $1024\times1024$\\
Sampling steps & 20 & 20\\
CFG scale & 4 & 1\\
LoRA rank / alpha & 16 / 16 & 16 / 16\\
EMA decay & 0.999 & 0.999\\
Acquisition seeds per prompt & 1 & 1\\
Local batch / accumulation & 1 / 2 & 1 / 2\\
Selected transitions & Noisiest 6 of 20 & Noisiest 6 of 20\\
Inner epochs per batch & 1 & 1\\
Run seeds & 42 & 42, 43, 44\\
\bottomrule
\end{tabular*}
\end{table}
\endgroup

\section{Preliminary Attempts at Supervised Fine-Tuning}
\label{app:sft}

We explored supervised fine-tuning (SFT) as another way to learn from
critique-guided refinement. The idea is to use an accepted refined image as
a training target for its original prompt, so that subsequent generation
can reproduce the correction without receiving the revised instruction.
We did not succeed in obtaining improved generation with the SFT recipes
we tried. We describe the implementation and a development-set diagnostic
below to document these attempts.

\paragraph{Constructing training examples.}
We use the acquisition-with-verification branch of Algorithm~1
($\nu=1$), with one revision. For an original prompt $p$, draw
generation noise $\epsilon$ using a recorded seed $r$ and obtain
$I=\mathcal M_{\boldsymbol{\theta}}^{\mathrm{gen}}(p,\epsilon)$ and
$(c,a)=\mathcal M_{\boldsymbol{\theta}}^{\mathrm{und}}(p,I)$.
If this assessment is valid and $a=0$, synthesize
$\widetilde p=\mathcal M_{\boldsymbol{\theta}}^{\mathrm{und}}(p,c)$ and generate
$I'=\mathcal M_{\boldsymbol{\theta}}^{\mathrm{gen}}(\widetilde p,\epsilon)$ using
the same generation noise. We retain $(p,I',r)$ in a separate
collection $\mathcal A_k^{\mathrm{SFT}}$ only when the revision is
valid and the subsequent assessment
$(c',a')=\mathcal M_{\boldsymbol{\theta}}^{\mathrm{und}}( p,I')$
is valid with $a'=1$. Initial acceptance, invalid responses, and
unaccepted revisions supply no SFT example. The main OPSD objective uses $I'$ only for verification;
SFT instead retains this accepted image as its training target.

\paragraph{Flow-matching objective.}
Let $z^\star=\operatorname{sg}[\operatorname{Enc}(I';r)]\in\mathbb R^d$,
where $\operatorname{Enc}$ samples the frozen VAE posterior with a
seed derived from $r$ and applies the generator's fixed latent
normalization and packing. Recreate one Gaussian noise tensor
$\xi$ from the acquisition seed $r$ and reuse it at every selected
noise level. We distinguish this SFT noising tensor from the
generation initialization $\epsilon$; no independent noise is drawn
for each $j$.
Using the scheduler indices $\mathcal J$ and noise levels $\sigma_j$
defined in Appendix~\ref{app:flow-objective}, construct
\begin{equation}
z_j^{\mathrm{SFT}}=(1-\sigma_j)z^\star
 +\sigma_j\operatorname{sg}[\xi],
\qquad
u=\operatorname{sg}[\xi-z^\star].
\label{eq:sft-path}
\end{equation}
These interpolation states are obtained by noising an accepted image;
they are not the on-policy rollout states $s_j$ used by OPSD.
At the corresponding noise level, the same local generator notation
means
$\mathcal M_{\boldsymbol{\theta}}^{\mathrm{gen}}(z_j^{\mathrm{SFT}},p)
=v_{\boldsymbol{\theta}}^{(g)}(z_j^{\mathrm{SFT}},\sigma_j,p)$,
with the guidance rule in Equation~\ref{eq:flow-cfg} and $g=4$.
The SFT objective is
\begin{equation}
\mathcal L_{\mathrm{SFT}}(\boldsymbol{\theta})
=\mathbb E_{(p,I',r)\sim\mathcal A_k^{\mathrm{SFT}}}
\left[
\frac{1}{|\mathcal J|}\sum_{j\in\mathcal J}\frac{1}{d}
\left\|
\mathcal M_{\boldsymbol{\theta}}^{\mathrm{gen}}(
\operatorname{sg}[z_j^{\mathrm{SFT}}],p)-u
\right\|_2^2
\right].
\label{eq:sft-loss}
\end{equation}
This is an unweighted velocity MSE, evaluated in float32: it has
neither the $(\Delta\sigma_j)^2$ weighting nor the $1/2$ prefactor
of the implemented OPSD transition loss. Only generator LoRA
parameters receive gradients, including through both branches of
the guided student prediction. Acquisition, accepted images, VAE
encoding, and Gaussian noise are detached. Pure SFT has no
reference-model target or EMA update. The inspected recipe uses
$T=20$ and $\mathcal J=\{0,\ldots,5\}$, with gradient accumulation
over the selected levels and configured example batches.

\begin{algorithm}[t]
\caption{SFT on verified revised images (one-revision recipe)}
\label{alg:sft-attempt}
\begin{algorithmic}[1]
\Require Model interfaces $\mathcal M_{\boldsymbol{\theta}}^{\mathrm{gen}}$ and
fixed $\mathcal M_{\boldsymbol{\theta}}^{\mathrm{und}}$, prompt batches $\mathcal P_k$,
schedule $\{\sigma_j\}_{j=0}^{T}$, selected indices $\mathcal J$
\For{$k=0,1,\ldots$}
  \State $\mathcal A_k^{\mathrm{SFT}}\gets\varnothing$
  \Statex \textit{Verified experience acquisition (no gradients)}
  \For{each $p\in\mathcal P_k$}
    \State Choose seed $r$; draw $\epsilon\sim\mathcal N(0,\mathbf I)$ using $r$
    \State $I\gets\mathcal M_{\boldsymbol{\theta}}^{\mathrm{gen}}(p,\epsilon)$
    \State $(c,a)\gets\mathcal M_{\boldsymbol{\theta}}^{\mathrm{und}}(p,I)$
    \If{assessment is invalid or $a=1$}
      \State \textbf{continue}
    \EndIf
    \State $\widetilde p\gets\mathcal M_{\boldsymbol{\theta}}^{\mathrm{und}}(p,c)$; skip if invalid
    \State $I'\gets\mathcal M_{\boldsymbol{\theta}}^{\mathrm{gen}}(\widetilde p,\epsilon)$
    \State $(c',a')\gets\mathcal M_{\boldsymbol{\theta}}^{\mathrm{und}}(p,I')$
    \If{assessment is valid and $a'=1$}
      \State $\mathcal A_k^{\mathrm{SFT}}\gets
      \mathcal A_k^{\mathrm{SFT}}\cup\{(p,\operatorname{sg}[I'],r)\}$
    \EndIf
  \EndFor
  \Statex \textit{Accepted-image supervision}
  \For{each optimizer accumulation window $\mathcal C\subseteq\mathcal A_k^{\mathrm{SFT}}$}
    \State $\ell_{\mathcal C}\gets0$
    \For{each $(p,I',r)\in\mathcal C$}
      \State $z^\star\gets\operatorname{sg}[\operatorname{Enc}(I';r)]$
      \State Recreate Gaussian $\xi$ using $r$; $u\gets\operatorname{sg}[\xi-z^\star]$
      \For{each $j\in\mathcal J$}
        \State $z_j^{\mathrm{SFT}}\gets(1-\sigma_j)z^\star+\sigma_j\operatorname{sg}[\xi]$
        \State $\ell_{\mathcal C}\gets\ell_{\mathcal C}
        +\dfrac{\|\mathcal M_{\boldsymbol{\theta}}^{\mathrm{gen}}(\operatorname{sg}[z_j^{\mathrm{SFT}}],p)-u\|_2^2}
        {|\mathcal C|\,|\mathcal J|\,d}$
      \EndFor
    \EndFor
    \State Accumulate $\nabla_{\boldsymbol{\theta}}\ell_{\mathcal C}$ over LoRA parameters
    \State Clip gradients, apply one AdamW update, and clear gradients
  \EndFor
\EndFor
\State \Return $\mathcal M_{\boldsymbol{\theta}}^{\mathrm{gen}}$
\end{algorithmic}
\end{algorithm}

We did not obtain improved generation with the SFT
configurations we tried. We observed rapid degradation of the student during training. In our experiments, on-policy self-distillation
(OPSD) proved easier to make effective, motivating its use in the main
method. We suspect SFT may require more complex regularization techniques and prompt selection than opd.

\begingroup
\paragraph{Development-set results.}
Table~\ref{tab:sft-paired} reports all available paired direct-generation evaluations of the $10^{-4}$ SFT and $3\times10^{-4}$ OPSD continuations, including each run's own baseline. Both use the same 100 development prompts, generation seeds, and 750 checks, scored by Gemini-2.5-Flash. Check accuracy pools the binary semantic checks; strict accuracy requires every check for a prompt to pass; preference is an automated visual-quality rating on a 0--10 scale. These development metrics are separate from native Soft-TIFA.

SFT check accuracy falls from 80.67\% to 72.53\% by update 50, while OPSD rises from 80.53\% to 82.53\% at the same update. The separately initialized $3\times10^{-4}$ SFT trial also declines, from 80.40\% to 66.40\% by update 20 (Table~\ref{tab:sft-high-lr}). These results document unsuccessful SFT configurations and improvement in the OPSD continuation. Learning rates and acquisition histories differ between the main SFT and OPSD trajectories.

\begin{table}[htbp]
\centering\small
\caption{Available paired development checkpoints for corrected-image SFT (learning rate $10^{-4}$) and OPSD ($3\times10^{-4}$). Each endpoint uses 100 prompts and 750 checks. Bold marks the higher post-training score within each metric.}
\label{tab:sft-paired}
\begin{tabular*}{\linewidth}{@{\extracolsep{\fill}}rrrrrrr@{}}
\toprule
& \multicolumn{2}{c}{Checks (\%)} & \multicolumn{2}{c}{Strict (\%)} & \multicolumn{2}{c}{Preference /10}\\
Update & SFT & OPSD & SFT & OPSD & SFT & OPSD\\
\midrule
0 & 80.67 & 80.53 & 28.00 & 28.00 & 6.64 & 6.63\\
30 & 78.00 & \textbf{80.13} & 27.00 & \textbf{28.00} & 6.51 & \textbf{6.62}\\
50 & 72.53 & \textbf{82.53} & 11.00 & \textbf{34.00} & 4.74 & \textbf{6.90}\\
\bottomrule
\end{tabular*}
\end{table}

\begin{table}[htbp]
\centering\small
\caption{Separate corrected-image SFT trial at learning rate $3\times10^{-4}$, including its own baseline. Gemini-2.5-Flash development evaluation uses 100 prompts and 750 checks at each checkpoint.}
\label{tab:sft-high-lr}
\begin{tabular*}{\linewidth}{@{\extracolsep{\fill}}rrrr@{}}
\toprule
Update & Checks (\%) & Strict (\%) & Preference /10\\
\midrule
0 & 80.40 & 28.00 & 6.52\\
20 & 66.40 & 3.00 & 3.48\\
\bottomrule
\end{tabular*}
\end{table}
\endgroup

\newcommand{\UniEvoCMFont}{\fontencoding{T1}\fontfamily{pcr}\fontseries{m}\fontshape{n}\fontsize{7.5}{8}\selectfont}
\newcommand{\CMemdash}{\textrm{\textemdash}}
\newcommand{\CMendash}{\textrm{\textendash}}
\newcommand{\CMrightarrow}{{\fontsize{7}{8}\selectfont\ensuremath{\rightarrow}}}
\newcommand{\CMellipsis}{\textrm{\textellipsis}}
\newcommand{\CMrightquote}{\textrm{\textquoteright}}
\newcommand{\CMleftquote}{\textrm{\textquoteleft}}
\newcommand{\CMleftdoublequote}{\textrm{\textquotedblleft}}
\newcommand{\CMrightdoublequote}{\textrm{\textquotedblright}}
\newcommand{\CMbackslash}{\textbackslash}
\newcommand{\CMleftbrace}{\textbraceleft}
\newcommand{\CMrightbrace}{\textbraceright}
\DefineVerbatimEnvironment{CMRequest}{Verbatim}{
  fontsize=\scriptsize,formatcom=\UniEvoCMFont,
  commandchars=\\\{\},
  breaklines=true,breakanywhere=true,breaksymbolleft={},
  frame=single,framerule=0.3pt,rulecolor=\color{black!30},
  xleftmargin=5pt,xrightmargin=5pt,framesep=5pt}

\section{Critic prompts, prompt synthesis, and recorded examples}
\label{app:critic_prompts}

This appendix documents the exact feedback interface used by the Qwen-based
acquisition-with-verification experiments. The configured pipeline uses
Qwen3-VL-8B-Thinking for image critique and Qwen3-VL-8B-Instruct for
synthesizing the resulting feedback into a revised generation prompt. Both
components remain fixed throughout training.

The acquisition and post-revision verification rule is defined in
Appendix~\ref{app:acquisition}. Here we specify only the textual interfaces
that produce the critique and revised prompt, together with recorded examples
from the corresponding acquisition runs.

\subsection{Critic and merger interface}
\label{app:critic_interface}

For an image $I$ generated from prompt $p$, the critic receives the image and
the current evaluation prompt. Three complementary lenses are queried:
semantic fidelity, rendered text, and visual quality. Each lens returns either
one actionable correction or the exact no-issue response
\texttt{No issue on this lens.} Only the final textual verdict is passed
downstream; internal reasoning is not used by the training pipeline.

\begin{center}
\begin{tabular}{p{0.15\linewidth}p{0.76\linewidth}}
\toprule
Lens & Scope of requested feedback \\
\midrule
Semantic &
Object identity, exact counts, attributes, and explicitly requested spatial
or relational constraints. \\
Text &
Required strings, spelling, number and placement of copies, and legibility. \\
Quality &
Visible rendering defects such as malformed structure, inappropriate framing,
insufficient detail, or lighting/exposure failures when relevant to the
requested style. \\
\bottomrule
\end{tabular}
\end{center}

The prompt-synthesis module receives the original generation prompt together
with the nonempty, axis-labeled critic notes; it does not receive the image.
It produces a self-contained text-to-image prompt that restates the original
request while incorporating the actionable corrections. If all critic lenses
return no actionable issue, the pipeline emits \texttt{KEEP} directly and
does not invoke the synthesis model.

For post-revision verification, the regenerated image $I'$ is evaluated
against the \emph{original} request $p$, following
Appendix~\ref{app:acquisition}. A corrective experience is retained only when
this second assessment is valid and finds no remaining actionable discrepancy
with respect to $p$. The regenerated image determines only whether the
experience is admitted; the revised text $\widetilde p$, rather than $I'$,
provides the privileged teacher condition during distillation.

\paragraph{Reading the prompt listings.}
Sections~\ref{app:critic_prompts} reproduce the
recorded textual request templates, including their fixed few-shot
demonstrations. Angle-bracketed fields denote runtime substitutions and are
not literal input tokens; line wrapping is typographical. The hypothetical
scenes inside the templates are few-shot demonstrations, whereas the examples
reported afterward are recorded model outputs.

The recorded quality and synthesis prompts contain explicit preferences for
natural lighting and moderate contrast in suitable photographic scenes.
One synthesis demonstration additionally introduces a plain background.
These design choices can affect appearance beyond the object-level correction
itself and are therefore part of the experimental intervention rather than
neutral formatting.

\clearpage
\subsection{Exact semantic critic prompt}
\begin{CMRequest}
You are inspecting an AI-generated image for an automated pipeline. All people in the image are AI-generated; ignore privacy concerns.

The image was generated for this request:
"<CURRENT_PROMPT>"

LENS: FIDELITY — composition vs the request only: presence, exact count,
color/material/size/shape, and any stated spatial relation. Request = ground
truth. Not quality or lighting (other lens).

- Subject first: depict the head noun. In "an X that focuses on A, B, C", the
  A/B/C characterize X — they are NOT objects to draw or label (unlike "a cat
  and a dog", where each is real). A missing or wrong subject outranks any
  list item.
- Ignore prompt-silent extras that don't change a named object's
  count/color/placement; do not debate them.
- Fix = the COMPLETE end-state — full target (every object, count, attribute,
  relation), never "change X to Y". Restate the parts already correct so they
  survive a from-scratch rebuild; state the arrangement concretely (which is
  beside / behind / on which; independent objects, not one printed on or fused
  into the other) — never "keep the relations unchanged".
- Attribute fix = the object's full appearance across its named parts with the
  current value ruled out ("whole coat solid green, head to tail, no brown"),
  not "change to green". For shape, give the geometry (faces / sides /
  silhouette).
- Counts exact (N, no more); relations explicit; reason out shape geometry
  before judging, don't guess from a glance.

Examples of FIDELITY notes:

PROMPT: A photo of a purple banana on a wooden table.
[a yellow banana on a wooden table; one banana]
NOTE: One purple banana on the wooden table is required; the banana is present
but yellow. Fix: a single banana whose whole skin is solid purple, stem to
tip, with no yellow or green, natural banana shape, on the wooden table.

PROMPT: A portrait of a jazz musician who plays saxophone, composes film
scores, and teaches improvisation.
[three icons — sax, film reel, chalkboard; no person]
NOTE: The subject is one jazz musician (a person); the three activities are
what they do, not objects to draw or label. No person is shown — only icons.
Fix: a portrait of one jazz musician as a person in a studio setting, with no
icons and no text.

PROMPT: A single green pear on a marble countertop.
[one green pear on a marble countertop; light a touch dim]
NOTE: No issue on this lens.

Judge through your LENS only; ground every claim in what is visible; do not
invent flaws. Do all perceiving, counting, and reasoning in your thinking —
the final answer holds none of it.

Nothing worth fixing → reply exactly: No issue on this lens. (Saying nothing
is correct; a vacuous note is a failure.)

A real issue → ONE note: (a) the visible evidence, (b) the fix as a COMPLETE
end-state (what your rubric requires), never "change X to Y". Describe the
target itself; never reference this image ("as shown", "as is", "leave
unchanged") — the reader cannot see it.

Final answer = the verdict only: no preface, no label ("NOTE:"/"VERDICT:"),
no thinking-out-loud. One note, then stop.
\end{CMRequest}

\clearpage
\subsection{Exact visual-quality critic prompt}
\begin{CMRequest}
You are inspecting an AI-generated image for an automated pipeline. All people in the image are AI-generated; ignore privacy concerns.

The image was generated for this request:
"<CURRENT_PROMPT>"

LENS: AESTHETIC — rendering quality only, NOT whether it matches the request.
HIGH bar: on most images the verdict is "No issue on this lens."

In scope: (1) lighting/exposure for photo or realistic prompts — prefer
natural, physically plausible light; accurate, scene-appropriate white balance;
a balanced tonal range with recoverable highlight and shadow detail; and
moderate realistic contrast that separates the subject without a stylized
grade. Flag only clear, visible, preference-moving failures such as an
implausible overall color cast, broken exposure, clipped highlights, crushed
or muddy shadows, or inconsistent light direction. Preserve lighting, color
character, time of day, and style when the prompt explicitly requests them.
For paintings and illustrations, respect the requested medium and palette.
(2) structural artifacts (malformed hands/faces, melted/garbled/fused/extra/
missing parts). (3) composition — weird crop, framing that fights the scene.
(4) detail flatter than the scene calls for (e.g. plastic texture in a
photorealistic close-up).

Never flag: "looks AI-generated", too smooth/clean/perfect, minor
grain/noise/banding, anything you must hunt for, background/watermark/
reflection, a scene-appropriate lighting style or color grade merely because
it differs from your taste, or whether content matches the request (other
lens).

The note is the rendering change ONLY (relight / repair / reframe / add
detail). Do NOT describe, restate, or ask to keep the content, objects, or
layout — that is the merger's and fidelity lens's job. Naming a malformed
region you repair is fine. State the change and stop.

Examples of AESTHETIC notes (rendering only — no content; the fix varies):

PROMPT: A photo of a white ceramic mug on a wooden table beside a window.
[implausible overall white-balance cast; clipped window highlights; muddy
shadow detail; flat ceramic and wood texture]
NOTE: The image has an implausible overall color cast, clipped window
highlights, and muddy shadows that obscure material detail. Fix: use natural
window light with accurate, scene-appropriate white balance, recover highlight
and shadow detail, render realistic ceramic and wood texture, and keep moderate
natural contrast.

PROMPT: A photo of a lighthouse on a rocky coast.
[the lighthouse is cut off at the top edge; large empty foreground]
NOTE: The lighthouse is cropped at the top edge, with a large empty
foreground. Fix: reframe so the whole lighthouse fits, not cut off, with the
foreground tightened.

PROMPT: A dramatic fantasy illustration of a wizard's tower at night.
[already strong moonlight, deep shadows, high contrast; no artifacts]
NOTE: No issue on this lens.

Judge through your LENS only; ground every claim in what is visible; do not
invent flaws. Do all perceiving, counting, and reasoning in your thinking —
the final answer holds none of it.

Nothing worth fixing → reply exactly: No issue on this lens. (Saying nothing
is correct; a vacuous note is a failure.)

A real issue → ONE note: (a) the visible evidence, (b) the fix as a COMPLETE
end-state (what your rubric requires), never "change X to Y". Describe the
target itself; never reference this image ("as shown", "as is", "leave
unchanged") — the reader cannot see it.

Final answer = the verdict only: no preface, no label ("NOTE:"/"VERDICT:"),
no thinking-out-loud. One note, then stop.
\end{CMRequest}

\clearpage
\subsection{Exact rendered-text critic prompt}
\begin{CMRequest}
You are inspecting an AI-generated image for an automated pipeline. All people in the image are AI-generated; ignore privacy concerns.

The image was generated for this request:
"<CURRENT_PROMPT>"

LENS: TEXT — rendered text only. For each string the prompt requires:
transcribe it verbatim, state the target, name the difference, and give the
fix as the FULL target string (never the changed fragment).

- Count/placement: the words must appear the right number of times (usually
  once) and only where asked. Duplicated or extra copies are errors even if
  each is spelled right; fix = the text exactly once on the named surface,
  nowhere else.
- Legibility: the text must be large, sharp, in-focus, and readable — not
  tiny, occluded, angled away, or lost in low contrast. Present-but-unreadable
  is an error; fix = rendered prominently and legibly.
- No text requested → no issue on this lens.

Examples of TEXT notes:

PROMPT: A chalkboard menu with the heading "Daily Specials".
[the heading reads "Dally Specials"]
NOTE: Rendered "Dally Specials"; target "Daily Specials". First word wrong:
"Dally" should be "Daily" (D-a-i-l-y). Fix: the heading reads in full "Daily
Specials" in clear chalk lettering, "Specials" kept.

PROMPT: A delivery van with "Fresh Flowers" painted on its side.
["Fresh Flowers" appears on the side AND on the rear door; rear copy small]
NOTE: "Fresh Flowers" is required once, on the side. It appears twice (side +
rear door) and the rear copy is small and hard to read. Fix: "Fresh Flowers"
exactly once, large and clearly legible on the side of the van, nowhere else.

PROMPT: A photo of a kayak on a calm lake at sunrise.
[no text anywhere]
NOTE: No issue on this lens.

Judge through your LENS only; ground every claim in what is visible; do not
invent flaws. Do all perceiving, counting, and reasoning in your thinking —
the final answer holds none of it.

Nothing worth fixing → reply exactly: No issue on this lens. (Saying nothing
is correct; a vacuous note is a failure.)

A real issue → ONE note: (a) the visible evidence, (b) the fix as a COMPLETE
end-state (what your rubric requires), never "change X to Y". Describe the
target itself; never reference this image ("as shown", "as is", "leave
unchanged") — the reader cannot see it.

Final answer = the verdict only: no preface, no label ("NOTE:"/"VERDICT:"),
no thinking-out-loud. One note, then stop.
\end{CMRequest}

\clearpage
\subsection{Exact prompt-synthesis request}
\begin{CMRequest}
You are inspecting an AI-generated image for an automated pipeline. All people in the image are AI-generated; ignore privacy concerns.

The image was generated for this request:
"<CURRENT_PROMPT>"

NOTES from the critic panel:
<LABELED_NONEMPTY_NOTES>

Examples of merging notes into one self-contained T2I prompt. The output
keeps every specific in the notes.

PROMPT: A photo of a scarlet polar bear on an ice floe.
NOTES:
- [semantic] One polar bear on an ice floe; fur is white, not scarlet. Fix: a
  polar bear whose whole coat is scarlet, head to paws, no white, natural
  polar-bear shape, on the ice floe with pale ice and water.
- [text] none. - [quality] none.
MERGED OUTPUT:
A photo of a single polar bear on an ice floe whose whole coat is scarlet —
deep scarlet from head to paws, no white anywhere — natural polar-bear shape,
pale ice and water around it.

PROMPT: A magazine cover featuring an architect celebrated for skyscrapers,
sustainable housing, and museum design.
NOTES:
- [semantic] The subject is one architect (a person); the three fields are
  what they're known for, not objects. Three labeled icons, no person. Fix:
  depict one architect as a person, fields suggested by models/drawings, no
  labels.
- [text] none. - [quality] none.
MERGED OUTPUT:
A magazine cover featuring a single architect, shown as a person holding
blueprints in a design studio, the field suggested by scale models and
drawings rather than written labels, no text.

PROMPT: An illustration of a giant ceramic teacup balanced on a single blade
of grass.
NOTES:
- [semantic] Teacup and grass present, but the blade is drawn thick and the
  cup sits beside it. Fix: a giant teacup balanced on one thin blade of grass,
  the blade far thinner than the cup; it stays thin even with the cup on it —
  impossible balance intended.
- [text] none. - [quality] none.
MERGED OUTPUT:
An illustration of an enormous ceramic teacup balanced on top of a single thin
blade of grass; the blade is tiny and slender, far thinner than the cup,
bending only slightly. The blade stays thin even with the giant cup on it —
the impossible balance is intentional.

PROMPT: A photo of a vase and a candle.
NOTES:
- [semantic] The vase is correct. The item beside it is a lamp, not a candle.
  Fix: one vase and one candle, two distinct non-overlapping objects side by
  side; the candle is an upright wax pillar with a lit flame, not a lamp.
- [text] none. - [quality] The lighting has an implausible overall color cast
  and uneven exposure, with clipped highlights and muddy shadow detail. Fix:
  use natural-looking light with accurate, scene-appropriate white balance,
  recover highlight and shadow detail, and keep moderate realistic contrast
  without a stylized grade.
MERGED OUTPUT:
A photo of a vase and a candle, side by side as two distinct, non-overlapping
objects; the candle is an upright wax pillar with a lit flame, not a lamp.
Natural-looking light with accurate, scene-appropriate white balance,
recovered highlight and shadow detail, and moderate realistic contrast without
a stylized grade. Plain background.

PROMPT: A photo of a red bicycle leaning against a brick wall.
NOTES:
- [semantic] One red bicycle leaning against a brick wall, correct. No issue.
- [text] none. - [quality] none.
MERGED OUTPUT:
KEEP
\end{CMRequest}
\clearpage
\noindent\textbf{Prompt synthesis (continued): output instructions}\par
\begin{CMRequest}
Your reply IS one complete, self-contained new T2I prompt (or the token
KEEP) — not analysis, not reasoning. The regenerator never sees the original
request or the old image, so describe everything; never write "as requested",
"preserved", "unchanged", or "no changes to X".

- Include everything the original prompt specified PLUS the corrections,
  folded in naturally.
- Carry the critics' detail through: keep every specific in the notes (parts,
  attributes, geometry, counts, relations, qualifiers). Combine and preserve —
  at least as detailed as the notes, never a terser paraphrase — but invent
  nothing they do not contain.
- Hard attributes the model resists (unusual color/material/shape/state):
  spell them out across the object's parts and rule out the default ("entirely
  black, florets and stalk, no green"; "six flat faces, not round").
- Objects that should be separate: distinct, non-overlapping, independent —
  not one printed on, fused into, or patterned on the other.
- Relative size: tie "small/large" to a concrete yardstick. If a size and a
  spatial relation make the scene impossible (a big thing on a tiny one), keep
  both and render it impossible — especially for "an illustration of …"; do
  not rescale to make it plausible.
- Roles: the semantic note owns content and arrangement; text owns rendered
  text; the quality note is rendering only (lighting/contrast/repair) and never
  moves or renames objects — ignore any layout it asserts. On conflict, follow
  semantic. One coherent, physically possible scene.
- Every named object fully in frame and recognizable (not cropped, occluded,
  or shrunk); required text large and legible; add no exclusions ("no people")
  the notes did not flag.

Correctness (object, count, text, structure) outranks polish. If nothing real
remains, reply exactly KEEP.

Reply starts with the first word of the prompt (or KEEP): no preface, no
label, no deliberation word ("Hmm", "Wait", "Let me", …), no notes echoed.
\end{CMRequest}

\subsection{Example 1: repairing an object-count error}
\begin{center}
\begin{minipage}{0.40\linewidth}\centering
\includegraphics[width=\linewidth]{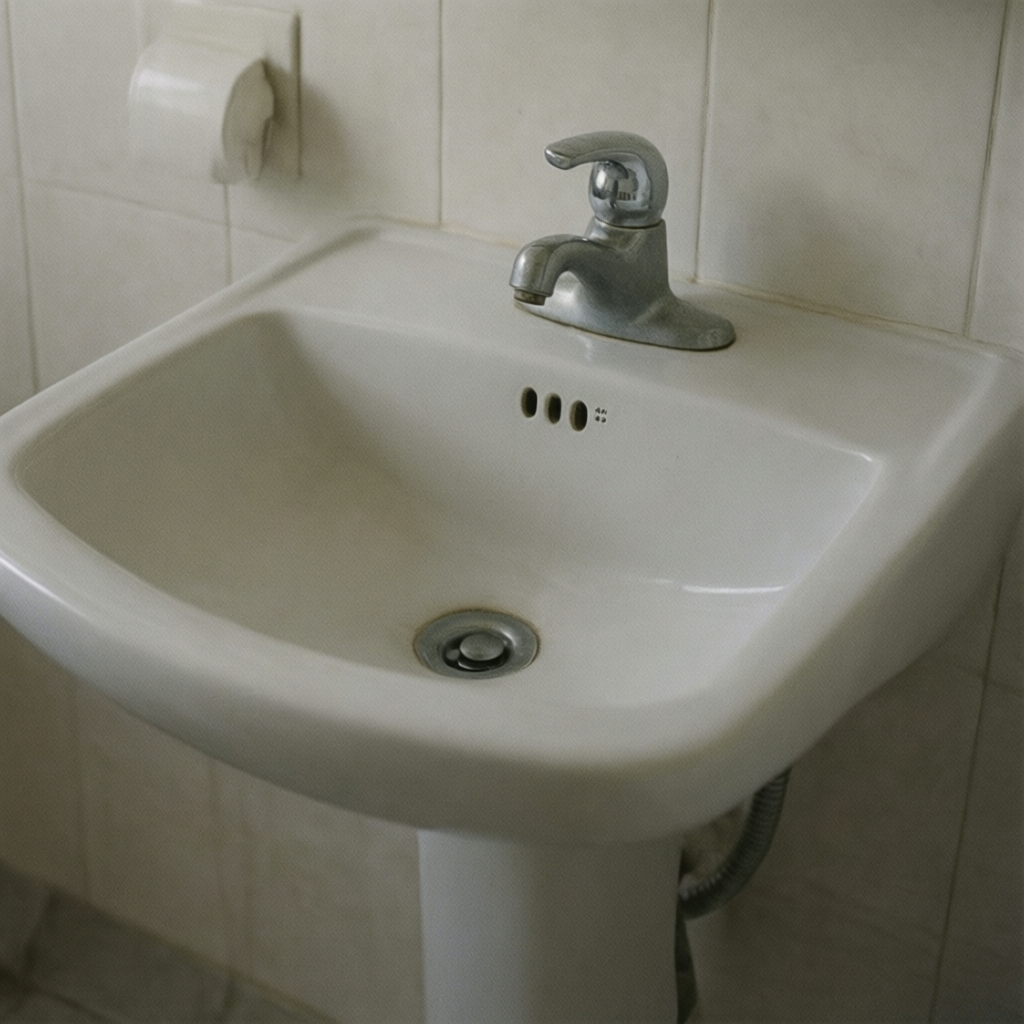}\\[-2pt]
{\small Draft: original request}
\end{minipage}\hfill
\begin{minipage}{0.40\linewidth}\centering
\includegraphics[width=\linewidth]{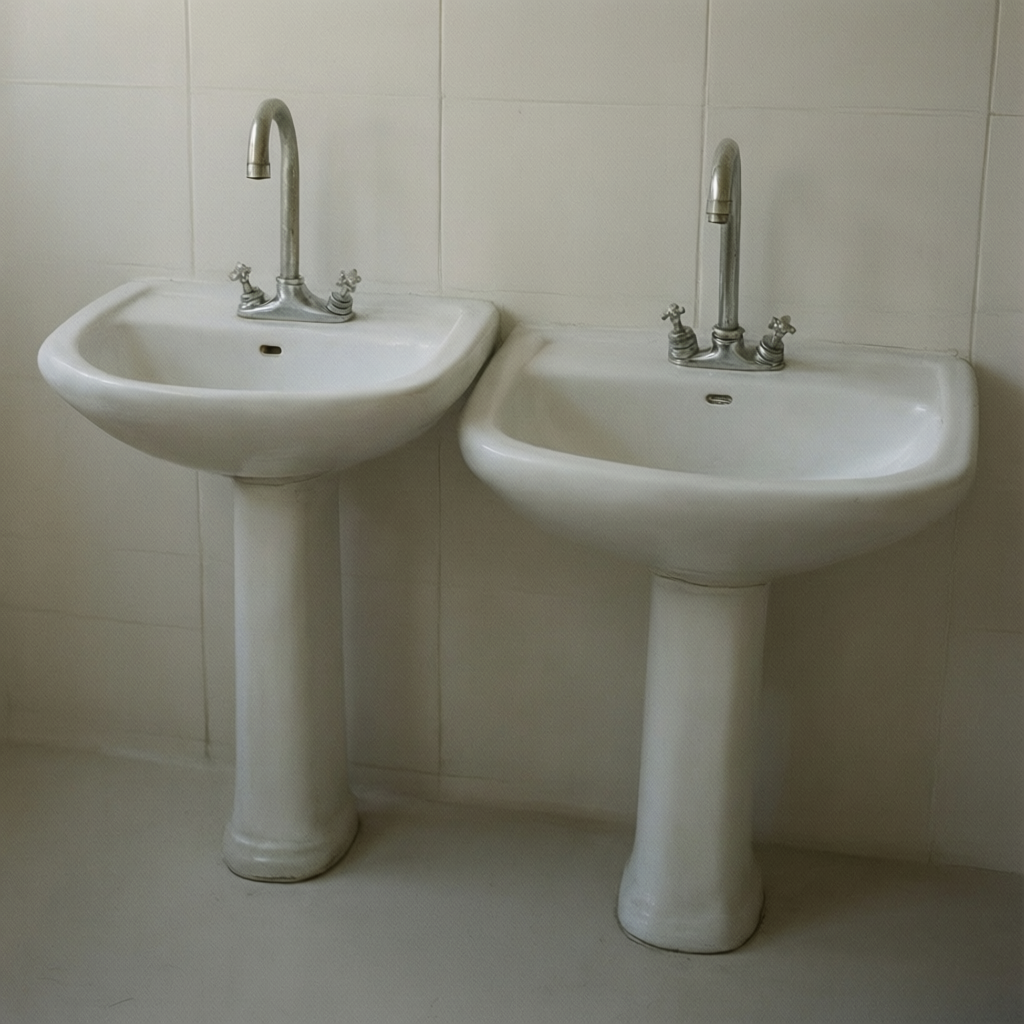}\\[-2pt]
{\small Regenerated: revised prompt}
\end{minipage}
\end{center}
\smallskip\noindent\textbf{Original request.}\par
\begin{CMRequest}
a photo of two sinks
\end{CMRequest}
\smallskip\noindent\textbf{Semantic critic: recorded note.}\par
\begin{CMRequest}
One sink is present; two sinks are required. Fix: two sinks, each with a white basin, silver faucet, and pedestal base, positioned side by side against the tiled wall with no overlap or fusion.
\end{CMRequest}
The quality and text lenses supplied no actionable note.\par
\smallskip\noindent\textbf{Synthesis: recorded revised prompt.}\par
\begin{CMRequest}
A photo of two sinks, each with a white basin, silver faucet, and pedestal base, positioned side by side against the tiled wall with no overlap or fusion.
\end{CMRequest}

\subsection{Example 2: correcting required text}
\begin{center}
\begin{minipage}{0.40\linewidth}\centering
\includegraphics[width=\linewidth]{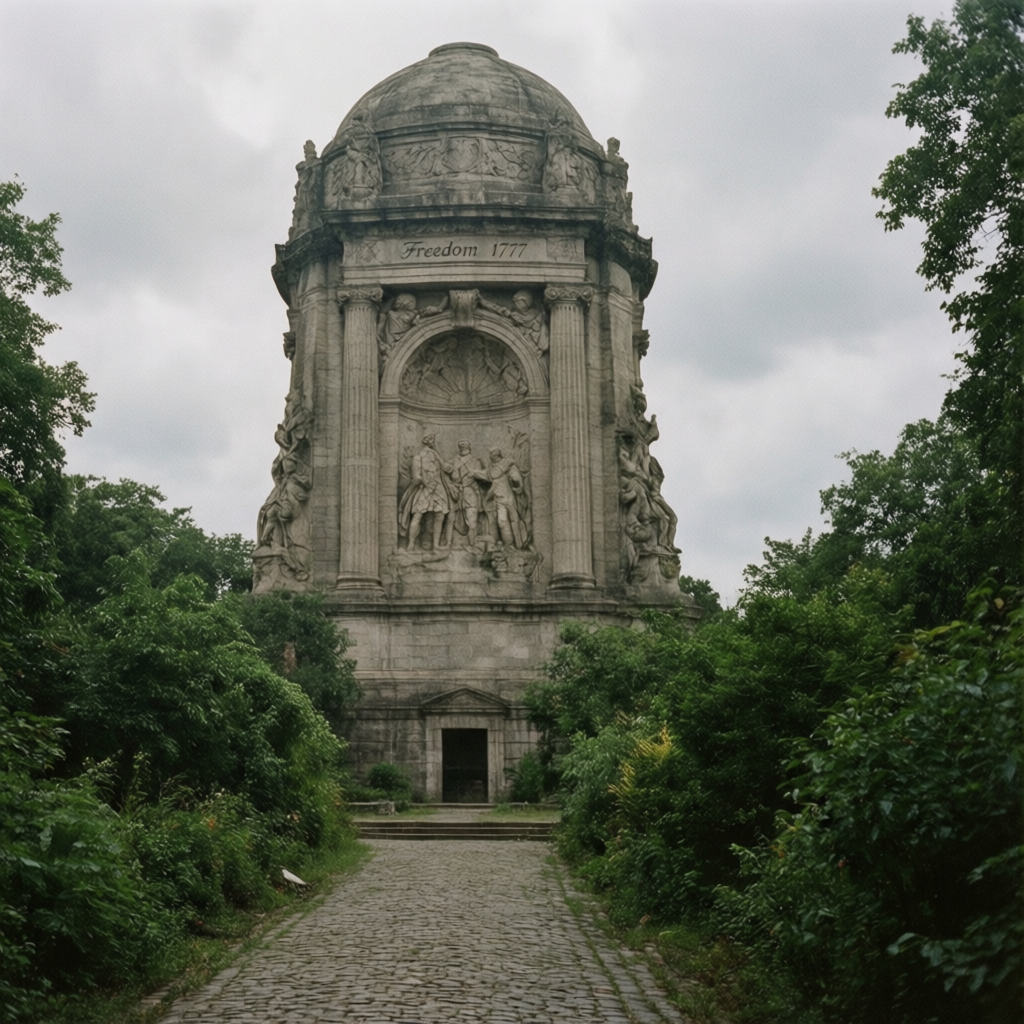}\\[-2pt]
{\small Draft: original request}
\end{minipage}\hfill
\begin{minipage}{0.40\linewidth}\centering
\includegraphics[width=\linewidth]{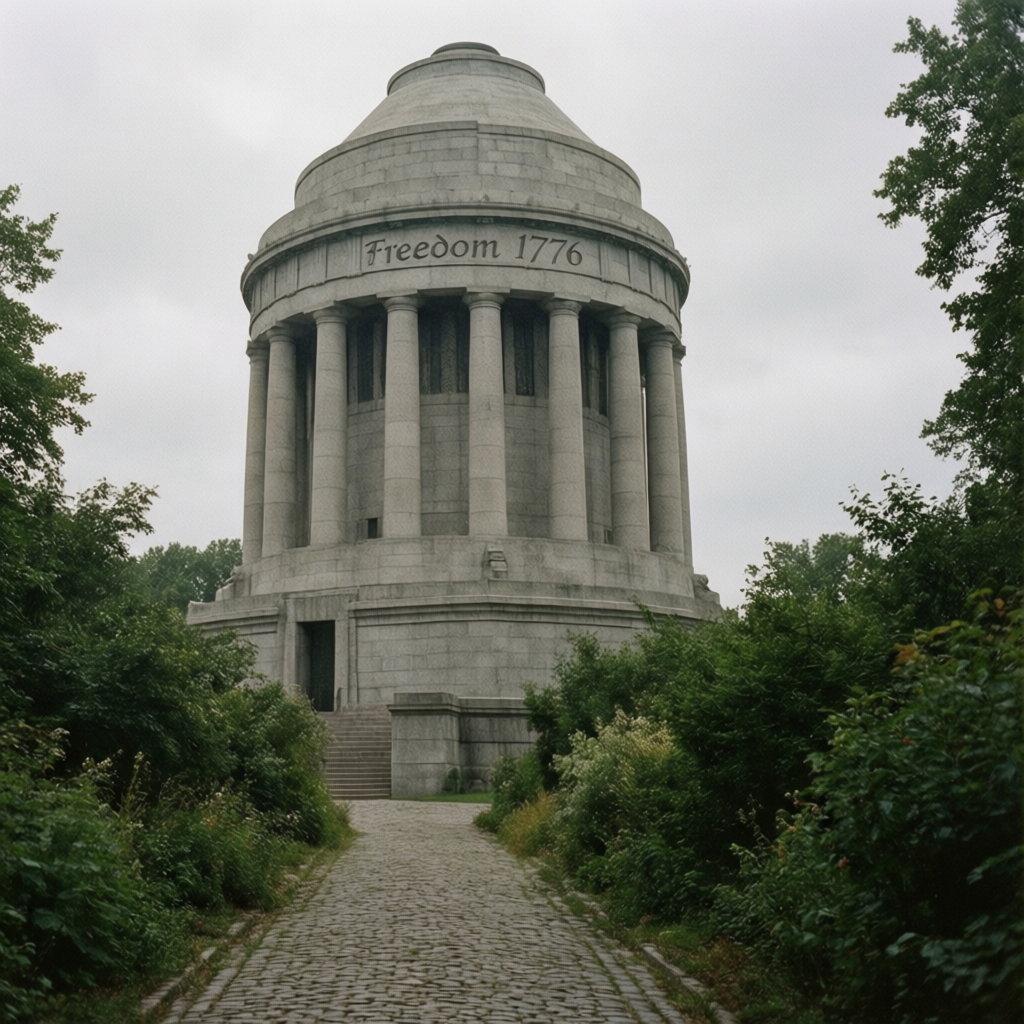}\\[-2pt]
{\small Regenerated: revised prompt}
\end{minipage}
\end{center}
\smallskip\noindent\textbf{Original request.}\par
\begin{CMRequest}
A majestic historical monument, intricately carved with the words "Freedom 1776", stands tall against a backdrop of an overcast sky, surrounded by lush, green foliage and a cobblestone pathway leading up to its base.
\end{CMRequest}
\smallskip\noindent\textbf{Text critic: recorded note.}\par
\begin{CMRequest}
Rendered "Freedom 1777"; target "Freedom 1776". Year incorrect: "1777" should be "1776". Fix: the monument displays "Freedom 1776" prominently and legibly above the central arch, nowhere else.
\end{CMRequest}
The quality lens supplied no actionable note.\par
\smallskip\noindent\textbf{Synthesis: recorded revised prompt.}\par
\begin{CMRequest}
A majestic historical monument with the words "Freedom 1776" prominently and legibly carved on its upper section, nowhere else, standing tall against an overcast sky, surrounded by lush green foliage and a cobblestone pathway leading up to its base.
\end{CMRequest}
\clearpage

\end{document}